 \documentclass[pdflatex,sn-mathphys-ay,iicol]{sn-jnl}

\usepackage{graphicx}%
\usepackage{multirow}%
\usepackage{amsmath,amssymb,amsfonts}%
\usepackage{amsthm}%
\usepackage{mathrsfs}%
\usepackage[title]{appendix}%
\usepackage{xcolor}%
\usepackage{textcomp}%
\usepackage{manyfoot}%
\usepackage{booktabs}%
\usepackage{makecell}%
\usepackage{algorithm}%
\usepackage{algorithmicx}%
\usepackage{algpseudocode}%
\usepackage{listings}%
\usepackage{soul}

\theoremstyle{thmstyleone}%
\theoremstyle{thmstyletwo}%

\theoremstyle{thmstylethree}%

\makeatletter
\AtBeginDocument{%
  \newif\ifNAT@yearlink@open
  \renewcommand*\hyper@natlinkstart[1]{\Hy@backout{#1}\global\NAT@yearlink@openfalse}%
  \renewcommand*\hyper@natlinkbreak[2]{#1\hyper@linkstart{cite}{cite.#2}\global\NAT@yearlink@opentrue}%
  \renewcommand*\hyper@natlinkend{\ifNAT@yearlink@open\hyper@linkend\global\NAT@yearlink@openfalse\fi}%
}
\makeatother

\makeatletter
\newcommand{\yrcite}[1]{(\hyper@linkstart{cite}{cite.#1}\citeyear{#1}\hyper@linkend)}
\makeatother

\makeatletter
\long\def\@tablecaption#1#2{%
  \setbox\tabcapbox\vbox{\tablecaptionfont%
  {\bfseries #1}{\hskip2mm}#2\vphantom{y}\par}%
  \box\tabcapbox%
}
\makeatother

\begin{document}

\title[Deep Registration of Drosophila Larval Brain Volumes]{Fast and Accurate Monomodal 3D High Resolution Deep Registration of Drosophila Larval Brain Volumes}


\author[1]{\fnm{Daniel} \sur{Reisenb\"uchler}}\email{daniel.reisenbuechler@informatik.uni-regensburg.de}
\author[1]{\fnm{Yousef} \sur{Sadegheih}}\email{yousef.sadegheih@informatik.uni-regensburg.de}
\author[1]{\fnm{Michael} \sur{Dittrich}}\email{michael.dittrich@informatik.uni-regensburg.de}
\author[1]{\fnm{Pratibha} \sur{Kumari}}\email{pratibha.kumari@informatik.uni-regensburg.de}
\author[1]{\fnm{Muhammad} \sur{Usman}}\email{muhammad.usman@informatik.uni-regensburg.de}
\author*[1,2]{\fnm{Dorit} \sur{Merhof}}\email{dorit.merhof@informatik.uni-regensburg.de}

\affil[1]{\orgdiv{Faculty of Informatics and Data Science}, \orgname{University of Regensburg}, 
\orgaddress{\city{Regensburg}, \country{Germany}}}

\affil[2]{\orgdiv{Fraunhofer Institute for Digital Medicine}, \orgname{MEVIS}, \orgaddress{\city{Bremen}, \country{Germany}}}


\abstract{The larval stage of \textit{Drosophila melanogaster} is a compact model system for neuroscience whose genetic toolkit allows fluorescent markers to be expressed in defined neural populations, and comparing the resulting expression patterns across animals requires every brain to be registered into a shared anatomical reference space. Existing pipelines for this task are predominantly based on classical registration methods,
which perform a new optimization for each volume, often require per-case parameter tuning,
and can take minutes per brain, limiting their use as a routine preprocessing step. We present a trained deep registration pipeline that deformably aligns a larval brain to a reference template in a single forward pass at high spatial resolution, on volumes that hold several times more voxels than those learned 3D registration is normally reported on, together with the preprocessing and anatomy-anchored evaluation pipeline required to apply it. Against eleven classical and seven further learned baselines on a held-out collection acquired with different acquisition and quality strata, the proposed pipeline is the most accurate, improving on the strongest classical baseline by 23 percentage points of anatomical landmark-local mutual information. It registers a volume one to two orders of magnitude faster than the classical deformable pipelines, and it retains more of its accuracy than any other method as acquisition quality degrades. The network, its trained weights and the full pipeline are released as the open-source deep larval brain registration framework: \url{https://github.com/agentdr1/deep-larval-brain-reg}.}





\maketitle

\section{Introduction}\label{sec:introduction}
The fruit fly \textit{Drosophila melanogaster} is a central model organism in neuroscience, supported by an extensive genetic toolkit for labeling, manipulating and tracing defined neural populations. At the larval stage, its brain contains only around $10,000$ neurons \citep{Nassif2002}, yet retains many structural features of the adult nervous system, making it a tractable system for studying neural organization at cellular resolution. Despite this anatomical simplicity, larvae exhibit a rich behavioral repertoire, enabling the study of sensory-driven behaviors, including responses to odors \citep{Gerber2006} and light \citep{Keene2012}. Genetic tools are central to mapping the larval brain. The GAL4/UAS system, for example, enables expression of markers such as green fluorescent protein within a defined subpopulation of neurons \citep{Brand1993,Venken2011}. Comparing these expression patterns across individuals requires a common reference space in which brains from different animals can be aligned. \cite{Muenzing2017} introduced a standard larval brain template derived from microscopy images that combine neuropil (NP, anti-N-cadherin antibody) and nerve tract (NT, anti-neuroglian antibody) staining. In this framework, anatomical NP channels from individual brains are registered to the template, and the resulting transformations are then applied to the corresponding NT and gene expression channels to bring them into the shared anatomical space.

The accuracy of this mapping depends critically on image registration. \cite{Muenzing2017} showed that an elastix-based sequential cascade (affine followed by B-spline) can produce high-quality larval brain alignments. However, four practical limitations restrict the applicability of classical registration as a routine preprocessing tool for large-scale studies. First, classical registration methods are slow: a single volume requires up to minutes of CPU optimization, and the most accurate deformable variants sit at the slow end of that range. This makes interactive use impractical and increases the cost of large-cohort studies. Second, new cases often require per-sample tuning of parameters such as step size, smoothing strength and related optimization settings. These settings must be retuned when acquisition conditions change, and the resulting configurations do not transfer cleanly across laboratories or microscope setups. Third, classical registration is optimized independently for each image pair and cannot learn from previous cases, which limits its ability to improve with additional data. Fourth, readily available registration methods for this domain are predominantly classical. No trained model for larval brain registration is available, so each new study assembles its own pipeline and repeats a per-volume optimization from scratch.

Learned registration instead trains a neural network to predict the deformation field from a moving
volume and a fixed template in a single forward pass. Once trained, inference is amortized into a single forward pass that takes well under a second per volume on a GPU, two orders of magnitude below the cost of the accurate classical methods. Moreover, the network requires a limited set of training-time hyperparameters, rather than a per-volume optimization schedule during inference. VoxelMorph \citep{voxelmorph} has become a reference framework for learned biomedical image registration, and subsequent backbones have extended this paradigm in several directions, including transformer- and attention-based architectures \citep{chen2022transmorph,chen2021vitvnet,sadegheih2024lhunet}, coarse-to-fine and recursive pyramids \citep{mok2020lapirn,wang2024rdp}, frequency-domain formulations \citep{jia2023fouriernet} and state-space models \citep{guo2024mambamorph}. We adopt several of these as learned baselines and describe each in Section~\ref{sec:methods}.

A further difficulty is the spatial resolution at which such a model has to operate. Many published studies of learned 3D deformable registration operate on volumes with roughly
$160$--$224$ voxels per dimension \citep{voxelmorph, chen2022transmorph, mok2020lapirn}, including the widely used $160 \times 192 \times 224$ brain volumes, whereas larval brain acquisitions retain
anatomical detail on a much finer in-plane grid. A $64 \times 768 \times 512$ volume holds about four times more voxels than a $160 \times 192 \times 224$ one, and a 3D convolutional network has to hold the activations of the entire volume in memory while it is trained. For the \textit{Drosophila melanogaster} larval brain it therefore remains unclear whether a learned model, under the architectural and memory constraints imposed by deep neural networks, can improve on classical pipelines at this resolution.

In this study, we present \textbf{D}eep \textbf{L}arval \textbf{B}rain \textbf{R}egistration (DLBR), a trained deep learning based registration pipeline for larval Drosophila brain volumes that is designed to remain robust to biological and experimental variation, including brain deformation, inhomogeneous staining and inconsistencies in sample preparation across data acquired over extended periods. We establish the necessary preprocessing steps, set up a comprehensive evaluation protocol that anchors intensity agreement to the annotated anatomical landmarks, and train a single network on the Janelia collection spanning different image-quality levels. We then evaluate DLBR against eleven classical and seven further learned registration methods on the independently curated Larvalign test set \citep{Muenzing2017}, which no method sees during training or model selection. DLBR is realized as the open-source deep-larval-brain-reg package, a computational framework that combines image preprocessing, network inference and evaluation into one ready-to-use pipeline for the registration of larval Drosophila brain microscopy images. Our contributions are summarized as follows.

\noindent\textbf{(i) High accuracy at high resolution:}
DLBR aligns larval Drosophila brain volumes to a common template at a resolution well above the one that learned 3D deep learning based registration studies commonly operate on, and is the most accurate of all classical and learned methods we compare.

\noindent\textbf{(ii) Fast inference:}
A single forward pass replaces the per-volume optimization of the classical pipelines, so the cost of optimization is paid once during training rather than again for every new brain. Registration drops from up to minutes to a fraction of a second per volume.

\noindent\textbf{(iii) Generalization and quality robustness:}
Trained on the Janelia collection alone and evaluated on the external Larvalign one, DLBR retains more of its accuracy than any other method we evaluate as acquisition quality degrades.

\noindent\textbf{(iv) Open-source deep-larval-brain-reg package:}
We release DLBR together with its trained weights in a PyTorch-based framework that supports preprocessing, inference and evaluation for larval Drosophila brain registration.

In the following, we introduce a comprehensive training and evaluation framework, and subsequently an extensive empirical evaluation across biological variation to demonstrate the practical applicability of the method for registration.

\section{Material}\label{sec:material}

\subsection{Larval Brain 3D Volume Data}
\begin{figure*}[tp]
\centering
\includegraphics[width=\textwidth,height=0.8\textheight,keepaspectratio]{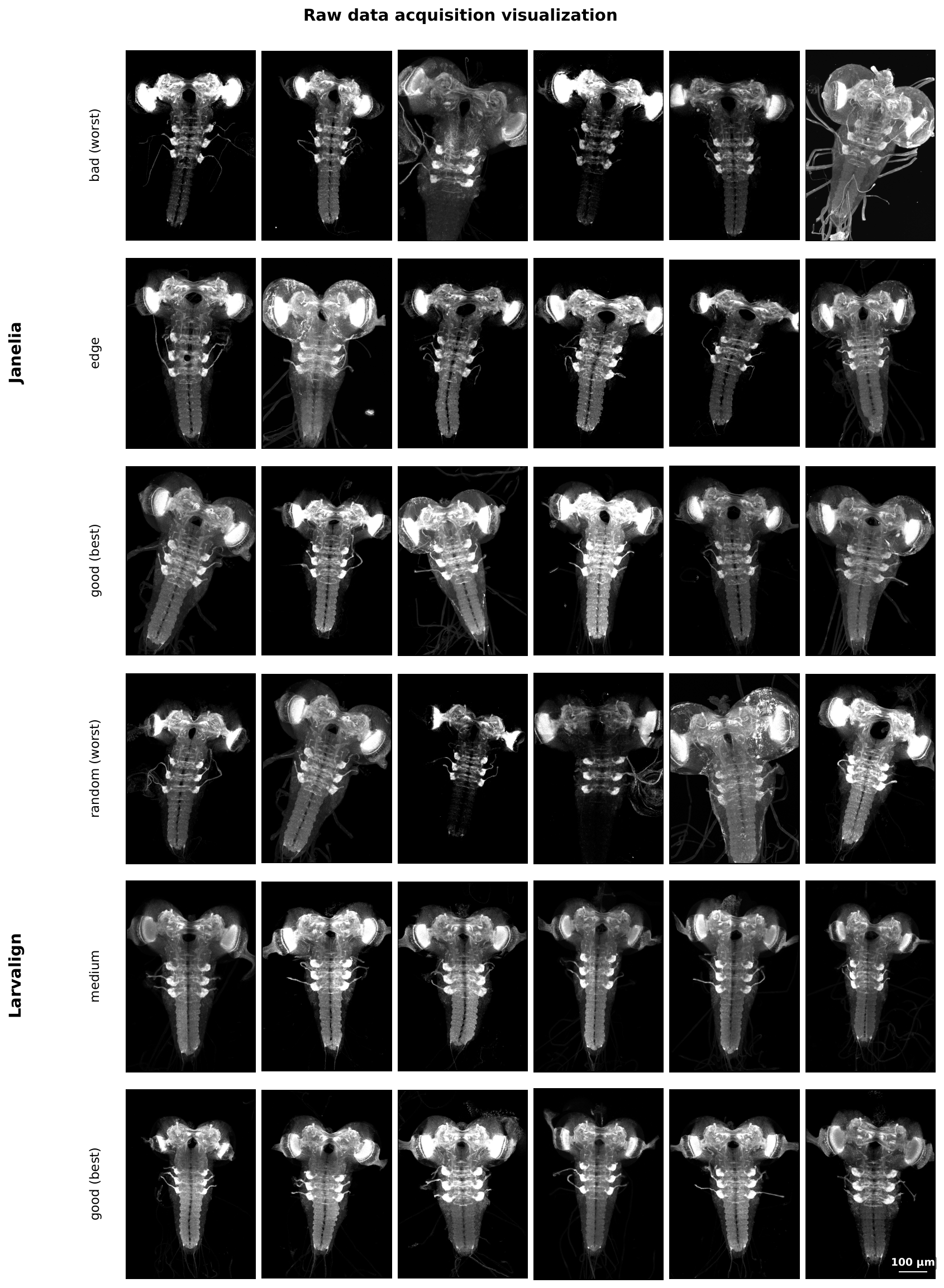}
\caption{\textbf{Example raw volumes from the Janelia and Larvalign datasets across image-quality strata.} Each panel is a $Z$ maximum-intensity projection, with rows for dataset and quality tier and columns for example specimens.}
\label{fig:raw_data_gallery}
\end{figure*}

We use two larval brain datasets. The first dataset, referred to here as Janelia, was acquired at the Janelia Research Campus (Virginia, USA). The second dataset is the Larvalign collection introduced by \cite{Muenzing2017}, an independently curated and image-quality stratified collection that we use only for testing. Sample sizes, native acquisition resolutions and image-quality stratifications for both datasets are summarized in Table~\ref{tab:data-overview}. Representative raw volumes from both datasets, spanning their different image-quality strata, are shown in Figure~\ref{fig:raw_data_gallery}. We maintained a strict separation between the two datasets: the Janelia data were used only for training and validation, whereas the entire Larvalign collection was held out as an external test set and was never used during training or model selection. The Janelia dataset was split into training and validation subsets using a validation fraction of $0.1$. The split was stratified by image quality so that the three quality strata (bad, edge, and good) were represented in the same $90/10$ proportion in both subsets, yielding $540$ training and $60$ validation volumes, or $180/20$ volumes per quality stratum. All reported test results were computed on the full Larvalign collection of $66$ volumes.

\begin{table}[t]
\centering
\footnotesize
\caption{\textbf{Overview of the two larval brain datasets used for training and external testing.} For each dataset, we report the native acquisition resolution, including voxel spacing (constant across volumes) and voxel-grid dimensions, together with the number of volumes in each image-quality stratum. The reported grid is the median across volumes, and the observed range along each of the $X$, $Y$ and $Z$ axes is listed beneath it.}

\label{tab:data-overview}
\setlength{\tabcolsep}{4pt}
\begin{tabular}{@{}ccccc@{}}
\toprule
\makecell{\textbf{Dataset}} & \makecell{\textbf{Spacing}\\(in $\mu$m)} & \makecell{\textbf{Voxel grid}\\($X{\times}Y{\times}Z$)} & \makecell{\textbf{Quality}} & \makecell{\textbf{\# Vol.}} \\
\midrule
\multirow{4}{*}{\shortstack{Janelia\\(Train/Val)}}   & \multirow{4}{*}{\shortstack[l]{$X$: $0.411$\\$Y$: $0.411$\\$Z$: $2.0$}}   & \multirow{4}{*}{\shortstack[l]{$978{\times}1430{\times}76$\\$X$: $890$--$1009$\\$Y$: $1256$--$1485$\\$Z$: $58$--$101$}}   & bad            & $200$          \\
                           &                                                    &                                                                                                                          & edge           & $200$          \\
                           &                                                    &                                                                                                                          & good           & $200$          \\
                           \cmidrule(l){4-5}
                           &                                                    &                                                                                                                          & \textbf{total} & $\mathbf{600}$ \\
\midrule
\multirow{4}{*}{\shortstack{Larvalign\\(Test)}} & \multirow{4}{*}{\shortstack[l]{$X$: $0.4566$\\$Y$: $0.4566$\\$Z$: $2.0$}} & \multirow{4}{*}{\shortstack[l]{$979{\times}1432{\times}79$\\$X$: $892$--$986$\\$Y$: $1258$--$1439$\\$Z$: $67$--$161$}} & random         & $25$           \\
                           &                                                    &                                                                                                                          & medium         & $21$           \\
                           &                                                    &                                                                                                                          & good           & $20$           \\
                           \cmidrule(l){4-5}
                           &                                                    &                                                                                                                          & \textbf{total} & $\mathbf{66}$  \\
\bottomrule
\end{tabular}
\end{table}

\subsection{Reference Template and Landmarks}
\label{subsec:template}
\begin{figure*}[t]
\centering
\includegraphics[width=\textwidth]{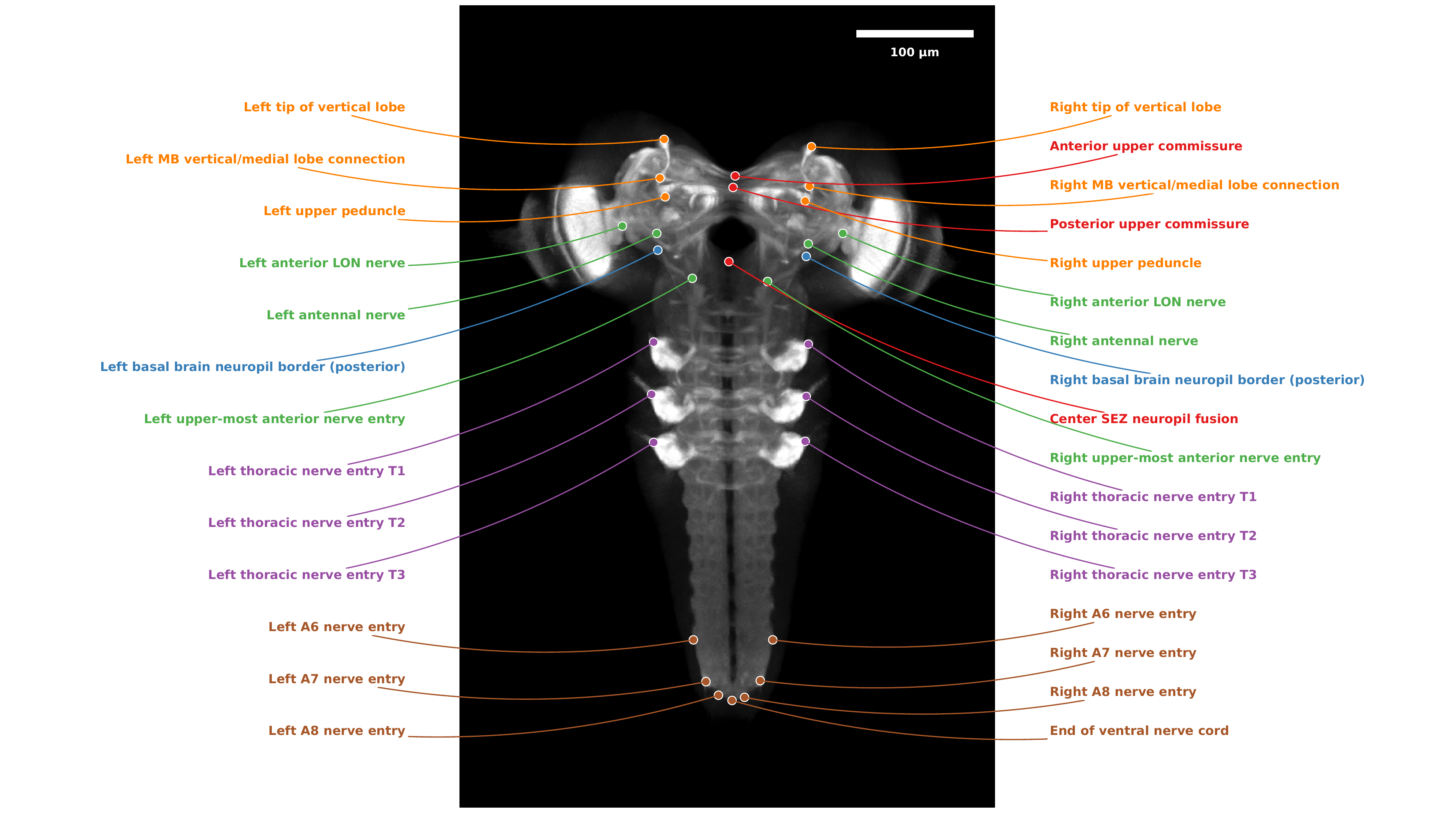}
\caption{\textbf{The anatomical landmarks on the Larvalign template.} Maximum-intensity projection with each landmark drawn as a labeled filled marker and colored by anatomical region. Landmarks are annotated only on the fixed template and are not available for a newly acquired volume.}
\label{fig:landmark_annotations}
\end{figure*}

We use a single fixed larval brain volume as the registration template, shared across all methods, scales and evaluations. This template is the Larvalign atlas of \cite{Muenzing2017}, on which we adopt the $L=30$ anatomical landmarks that were carefully annotated by domain experts in neurobiology for the Larvalign study. The landmarks cover the principal anatomical structures, including the anterior and posterior upper commissure, the central subesophageal zone, the mushroom-body vertical and medial lobes, the peduncles, the antennal and anterior larval optic nerves, the upper-most anterior nerve entries, the posterior basal brain neuropil borders, the paired thoracic nerve entries ($T_1$--$T_3$), the abdominal nerve entries ($A_6$--$A_8$), and the posterior tip of the ventral nerve cord. Their spatial layout on the template is shown in Figure~\ref{fig:landmark_annotations}, where each landmark is labeled with its full anatomical name and color-coded by anatomical region. These landmarks are annotated only on the template volume. They are propagated into the evaluation metrics by sampling local neighborhoods on the fixed (template) side, as described in Section~\ref{subsec:metrics}. This removes the need for per-volume manual annotation of the moving images. The shared reference template and its $30$ landmarks are shown in Figure~\ref{fig:preprocessing}(c), next to a newly acquired case before and after preprocessing in panels (a) and (b).

\section{Methods}\label{sec:methods}

\subsection{Preprocessing}
\label{subsec:preprocessing}

\begin{figure}[]
\centering
\includegraphics[width=\columnwidth]{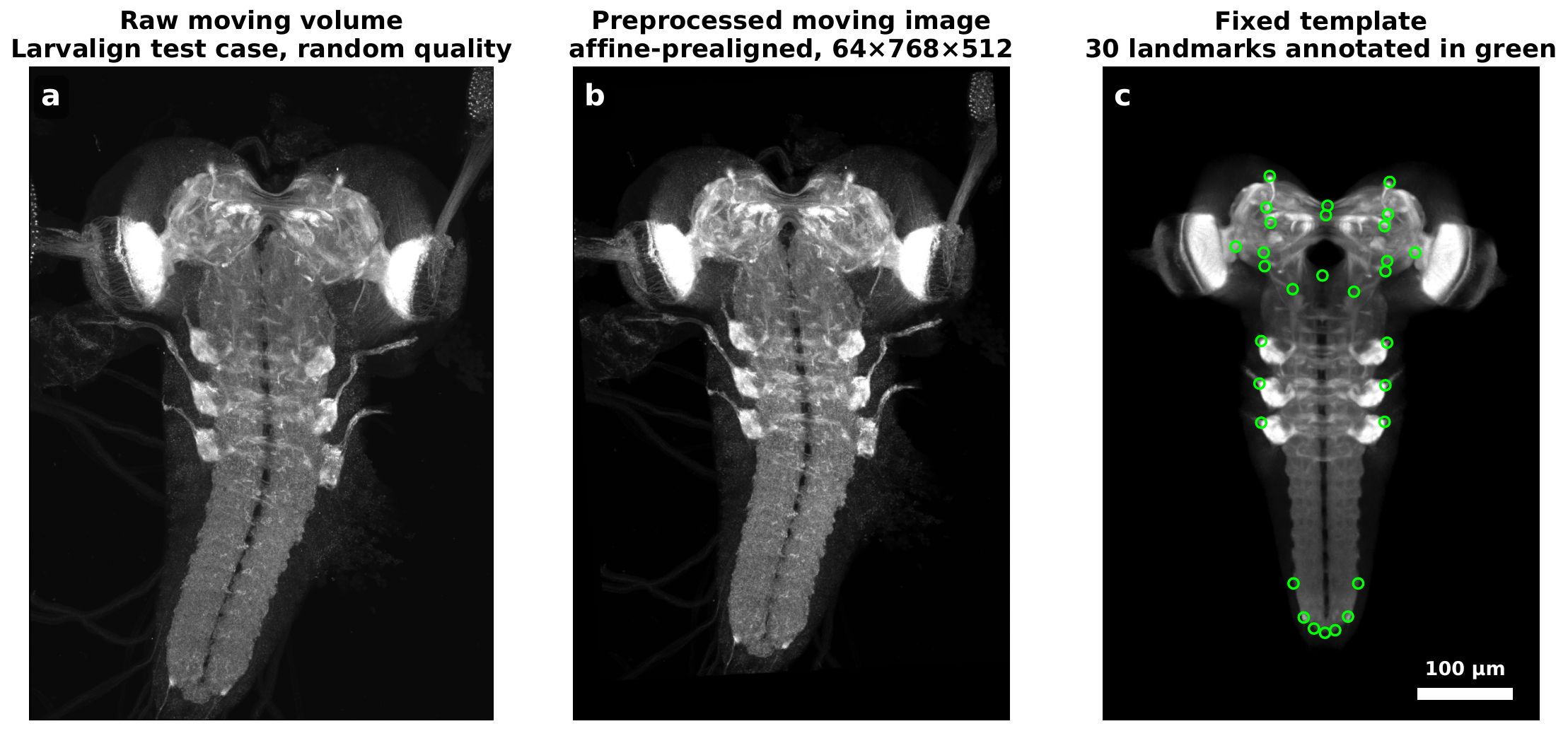}
\caption{\textbf{Preprocessing pipeline and reference template.} Maximum-intensity $Z$
projections. (a) A raw Larvalign acquisition at native resolution. (b) The same volume
after affine pre-alignment and resampling onto the fixed $64\times768\times512$ grid, which
is the input the model consumes. (c) The atlas template with the $L=30$ landmarks
overlaid.}
\label{fig:preprocessing}
\end{figure}

All raw volumes passed through a single preprocessing pipeline, applied
identically to the Janelia and Larvalign collections so that every downstream method operates on the same template pre-aligned, intensity-normalized
volumes (Figure~\ref{fig:preprocessing}). Janelia volumes were stored as compressed Zeiss LSM stacks and
Larvalign volumes as multi-page TIFFs. We transposed each volume to a canonical $(Z,Y,X)$ axis order with the sectioning axis as $Z$. Each volume was then intensity-normalized by clipping to the $0.01$ and
$99.9$ percentiles of its foreground voxels and rescaling to $[0,1]$ in
single precision. Restricting the percentiles to foreground voxels
prevents the large background region from dominating the statistics, and
the clipping suppresses saturated outliers while preserving intra-brain
contrast. The normalized volume was registered to the shared template with an affine transform, using SimpleITK's multi-resolution mutual-information framework, following \cite{mattes2001nonrigid,mattes2003petct,lowekamp2013simpleitk}. We used a Mattes
mutual-information metric with $50$ histogram bins and $0.2\%$ seeded
random voxel sampling, geometric initialization of the transform, and
regular-step gradient-descent optimization over a four-level image
pyramid.
This affine stage removed coarse differences in pose, scale and orientation so
that all methods began from a common baseline, and the \texttt{identity}
comparator as naive baseline measures how much alignment this step alone achieves. Finally, each aligned volume was resampled onto a fixed reference grid by cubic B-spline interpolation, preserving the template's physical extent and clipping any interpolation overshoot back to $[0,1]$. We generated
three target grids that span a geometric progression in in-plane
resolution while keeping the anisotropic sectioning axis fixed at $Z=64$,
$(Z,Y,X)\in\{(64,256,128),\,(64,512,256),\,(64,768,512)\}$. The highest-resolution grid $(64,768,512)$ is the featured
scale used in the main results, with the two coarser grids supporting the
resolution-scale ablation of Section~\ref{subsec:ablation}. Figure~\ref{fig:preprocessing}(b) shows a representative volume after this affine pre-alignment. Further examples spanning the acquisition-quality tiers of the Larvalign test set are shown in Figure~\ref{fig:qualitative_overlay}, and the corresponding raw volumes of both datasets in Figure~\ref{fig:raw_data_gallery}.

\subsection{3D Image Registration}
\label{subsec:3d-image-registration}

Let $\Omega\subset\mathbb{R}^3$ denote the image domain, i.e.\ the spatial
extent of the volume grid. Registration always acts on a pair of volumes. Throughout
this paper the fixed image $I_f$ is the shared reference template of
Section~\ref{subsec:template}. It is the same volume for every case and defines the
anatomical space into which all brains are mapped. The moving image $I_m$ is a newly
acquired brain volume that has to be brought into that space. A registration method
estimates a spatial transformation $\phi$ that relates positions in the fixed image to
positions in the moving image, and applying it to $I_m$ produces the moved image
$I_m\circ\phi$, which is the new acquisition resampled onto the template grid.
Registration quality is then judged by how closely the moved image agrees with the fixed
template.

Classical deformable image
registration is formulated as the search for a
spatial transformation $\phi:\Omega\to\Omega$ that maximizes a similarity
measure $\mathcal{S}(I_f, I_m\circ\phi)$ between a fixed image $I_f$ and
a moving image $I_m$ warped by $\phi$, subject to a smoothness
prior $\mathcal{R}(\phi)$. The transformation is typically composed of a
linear pre-alignment (e.g., affine) followed by a non-linear
component, parameterized either as a free-form B-spline deformation, introduced by \cite{Rueckert1999}, or as a velocity field whose flow defines a
diffeomorphism, as in \cite{voxelmorph}. Two families of methods dominate the practical landscape and are used as baselines throughout this paper: Classical registration methods optimize each image pair independently, whereas deep-learning-based methods learn the registration during training and apply it directly to new cases at inference.

\subsection{Classical Registration Methods}
\label{subsec:classical-3d-image-registration}
Classical methods optimize the transformation $\phi$ separately for each
moving--fixed pair, with no information shared across cases. We compare
against eleven such baselines, all evaluated on the same preprocessed,
template pre-aligned volumes and scored under the same metrics as the
learned model. The \texttt{identity} comparator applies no deformation
beyond the preprocessing affine and serves as a reference floor that shows
how much of the alignment is already achieved before any method is run.
As linear baselines we use rigid and affine
registration from the SimpleITK toolkit of \cite{lowekamp2013simpleitk},
each driven by a Mattes mutual-information
metric, following \cite{mattes2003petct}, optimized by regular-step gradient descent
over a four-level image pyramid. The same two transform models are also run through
\texttt{elastix}, following \cite{klein2010elastix}, which brings its own default metric
parameterization and a stochastic optimizer, so each linear transform model is
represented by two independent implementations rather than by a single toolkit. The deformable baselines are built around the
free-form B-spline deformation of \cite{Rueckert1999}, in which a lattice
of control points parameterizes a smooth
displacement field. We evaluate a bare SimpleITK B-spline, an
\texttt{elastix} B-spline that adds an explicit bending-energy penalty for
a smoother field, a two-stage \texttt{elastix}
pipeline that runs an affine map followed by a B-spline, and the NiftyReg
\texttt{reg\_f3d} free-form deformation of \cite{modat2010niftyreg} as a third independent B-spline
baseline. These four share one transform model but differ in how they constrain it. The
SimpleITK variant is driven by a quasi-Newton optimizer and is regularized only through
the coarseness of its control-point lattice, the \texttt{elastix} variants add a
bending-energy penalty under a stochastic optimizer, and NiftyReg carries an explicit
bending-energy weight. The free-form deformation is therefore not judged through a single
implementation of it. Two diffeomorphic baselines enforce
invertibility by construction: SimpleITK diffeomorphic
Demons of \cite{vercauteren2009demons}, which
Gaussian-smooths the deformation at every iteration, and ANTs symmetric
normalization (SyN) of \cite{avants2008syn}, the de facto gold-standard classical
deformable method, run in its deformable-only mode because the data are
already affine-aligned. Every method is otherwise run under the defaults of its own
toolkit, including its similarity metric, its multi-resolution scheme and its convergence
criteria.

\subsection{Deep Learning based Registration with DLBR}
\label{subsec:deep-learning-based-registration}
\begin{figure*}[t]
\centering
\includegraphics[width=\textwidth]{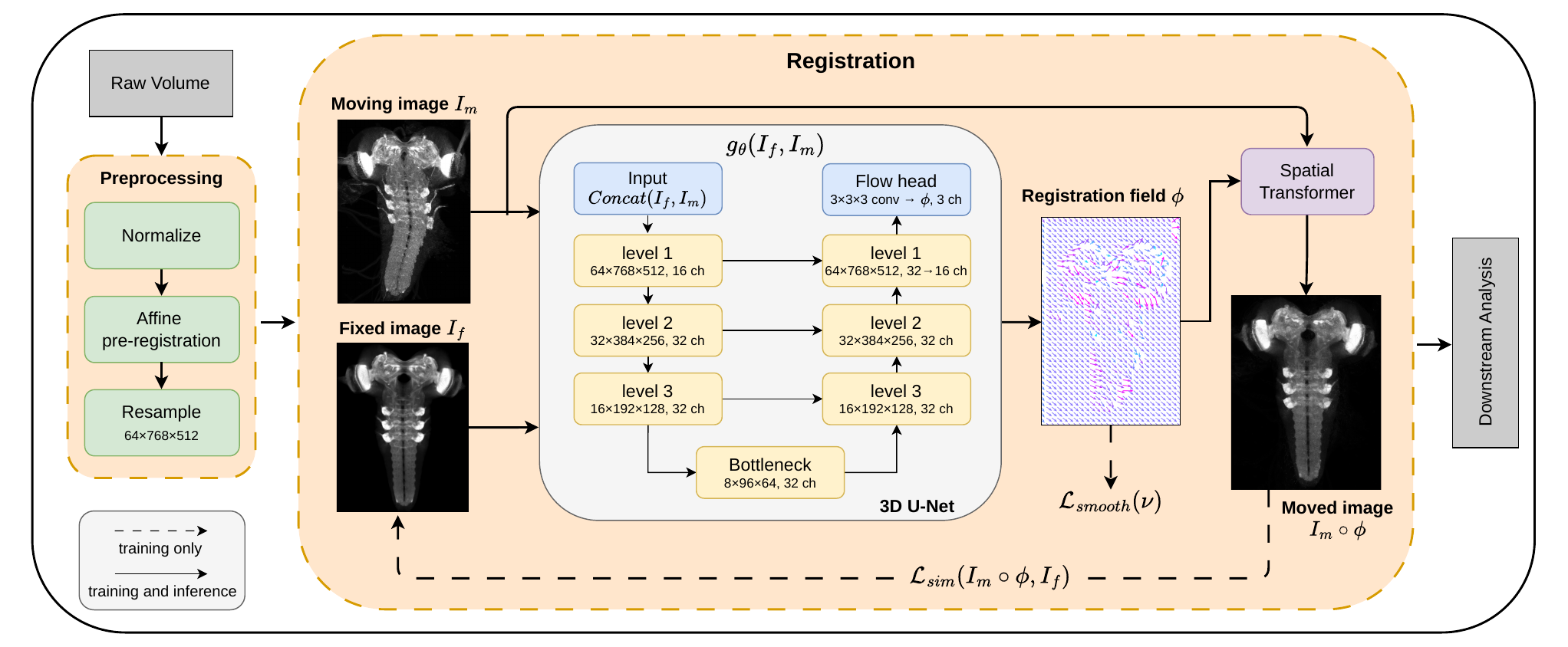}
\caption{\textbf{Overview of the DLBR registration workflow.} Preprocessing normalizes a
raw acquisition, pre-registers it affinely to the template and resamples it onto the fixed
$64\times768\times512$ grid. The resulting moving volume $I_m$ and the fixed template
$I_f$ are concatenated into a two-channel input and passed through the 3D U-Net
$g_\theta$, with the resolution and channel width of every encoder, bottleneck and decoder
level annotated in the figure. A final $3\times3\times3$ convolution emits the
three-channel registration field $\phi$, and a spatial transformer warps $I_m$ into the
moved volume $I_m\circ\phi$, which is what downstream analysis consumes. The similarity
loss between moved volume and template and the smoothness penalty on the predicted field
are evaluated during training only, while every other stage runs at both training and
inference time.}
\label{fig:method-overview}
\end{figure*}

DLBR is the registration pipeline we propose. It couples the preprocessing of
Section~\ref{subsec:preprocessing} to a learned registration model, and that model follows
the VoxelMorph approach of \cite{voxelmorph}, which replaces the per-sample optimization
of the classical methods with a
single convolutional network $g_\theta$ that learns to register any moving
volume $I_m$ to the shared template $I_f$. The network is trained once over
the Janelia collection and, at test time, predicts a dense deformation
$\phi$ for an unseen volume in a single forward pass, so the cost of
optimization is paid during training rather than separately for every new
brain. The remainder of this section describes the network, its training and the
evaluation protocol in turn, and Figure~\ref{fig:method-overview} gives an overview of the
full pipeline from a raw acquisition to an aligned volume.

\subsubsection{Network architecture}

The moving and fixed volumes are stacked into a two-channel input
$[I_m,I_f]\in\mathbb{R}^{2\times D\times H\times W}$ and passed through the
3D U-Net of \cite{ronneberger2015unet}. Both volumes enter the network, because a
displacement can only be predicted from the difference between where anatomy currently
sits in the moving volume and where the template expects it. The fixed template is used a
second time at training time, where it forms one side of the similarity term.
The encoder has four levels. Each
level applies two consecutive $3\times3\times3$ convolutions, each followed by
a LeakyReLU activation with negative slope $0.2$, and the three transitions
between levels halve the spatial resolution through $3\times3\times3$
convolutions of stride two. Encoder feature widths are $16$, $32$, $32$ and
$32$ channels, so the bottleneck sits at one eighth of the input resolution
along every axis, which is a grid of $8\times96\times64$ voxels at the featured
scale.

The decoder mirrors this contracting path in two phases. Three upsampling
stages restore resolution by trilinear interpolation to the size of the
matching encoder level, concatenate that level through a skip connection and
apply a further pair of convolutions, with widths $32$, $32$ and $32$. Four
refinement blocks then operate at full resolution with widths $32$, $32$, $16$
and $16$, again as convolution pairs. The skip connections let the coarse pose
of the brain and the fine detail around individual structures be recovered
together. A final $3\times3\times3$ convolution maps the $16$ decoder features
to a three-component field $u:\Omega\to\mathbb{R}^3$ over the image grid.
Table~\ref{tab:architecture} summarizes the
resolution and width at every stage. The complete network holds $604{,}195$
trainable parameters. The network is therefore small, but a single full-resolution
feature map at the featured grid already holds more than $400$ million values, so the
memory needed to train it is set by the activations rather than by the weights.

\begin{table}[t]
\centering
\footnotesize
\caption{\textbf{DLBR 3D U-Net configuration.} Every encoder and decoder block is a pair of $3\times3\times3$ convolutions with LeakyReLU activations of negative slope $0.2$. Downsampling uses strided convolutions, and each upsampling stage interpolates to the size of the matching encoder level and concatenates it through a skip connection.}
\label{tab:architecture}
\begin{tabular}{@{}llcc@{}}
\toprule
\textbf{Stage} & \textbf{Block} & \textbf{Resolution} & \textbf{Ch.} \\
\midrule
Input & concat $[I_m,I_f]$ & $64{\times}768{\times}512$ & $2$ \\
\midrule
\multirow{4}{*}{Encoder} & level 1 & $64{\times}768{\times}512$ & $16$ \\
 & level 2 & $32{\times}384{\times}256$ & $32$ \\
 & level 3 & $16{\times}192{\times}128$ & $32$ \\
 & bottleneck & $8{\times}96{\times}64$ & $32$ \\
\midrule
\multirow{7}{*}{Decoder} & upsample 1 + skip & $16{\times}192{\times}128$ & $32$ \\
 & upsample 2 + skip & $32{\times}384{\times}256$ & $32$ \\
 & upsample 3 + skip & $64{\times}768{\times}512$ & $32$ \\
 & refine 1 & $64{\times}768{\times}512$ & $32$ \\
 & refine 2 & $64{\times}768{\times}512$ & $32$ \\
 & refine 3 & $64{\times}768{\times}512$ & $16$ \\
 & refine 4 & $64{\times}768{\times}512$ & $16$ \\
\midrule
Flow head & $3{\times}3{\times}3$ conv & $64{\times}768{\times}512$ & $3$ \\
\bottomrule
\end{tabular}
\end{table}

\subsubsection{Deformation model and spatial transformer}

The predicted field is a dense displacement in voxel units, ordered
$u=(u_z,u_y,u_x)$ to match the canonical $(Z,Y,X)$ axis order of the
preprocessed volumes. It defines the deformation
\begin{equation}
    \phi(p) \;=\; p + u(p), \qquad p\in\Omega,
    \label{eq:deformation}
\end{equation}
so that the moved volume is obtained by sampling the moving volume at the
displaced positions,
\begin{equation}
    (I_m\circ\phi)(p) \;=\; I_m\bigl(p+u(p)\bigr).
    \label{eq:warp}
\end{equation}
Because $p+u(p)$ is in general not a grid point, the right hand side is
evaluated by trilinear interpolation over the eight neighboring voxels. This
makes the warp differentiable with respect to both the displacement and the
intensities of $I_m$, so gradients propagate back to the network weights
$\theta$. We implement it as the differentiable spatial transformer of
\cite{jaderberg2015spatial}. Sampling positions are expressed in normalized
coordinates on $[-1,1]$ along each axis, so a displacement of one voxel along
an axis of length $n$ corresponds to a normalized step of $2/(n-1)$, with the
outermost voxel centers placed exactly at $-1$ and $1$. Positions that fall
outside the volume are resolved by clamping to the border rather than by zero
padding, which prevents the warp from introducing spurious dark tissue where a
brain extends beyond the field of view.

\subsubsection{Training objective}

Training is fully unsupervised and requires no landmarks or segmentations
on the moving volumes. The total objective combines an image-similarity
term on the moved-and-fixed pair with a smoothness penalty on the field
predicted by the network,
\begin{equation}
\begin{split}
    \mathcal{L}(\theta;\,I_m,I_f)
    &\;=\; \mathcal{L}_{\mathrm{sim}}\!\bigl(I_m\circ\phi,\,I_f\bigr) \\
    &\quad+\; \lambda_{\mathrm{smooth}}\,
              \mathcal{L}_{\mathrm{smooth}}(u).
\end{split}
\label{eq:total-loss}
\end{equation}
For the similarity term we use normalized cross-correlation (NCC), computed over the
whole volume,
\begin{equation}
    \mathcal{L}_{\mathrm{NCC}}(I, J)
    \;=\;
    1 \;-\;
    \frac{\sum_{p}\tilde I(p)\,\tilde J(p)}
         {\sqrt{\sum_{p}\tilde I(p)^{2}}
          \sqrt{\sum_{p}\tilde J(p)^{2}}+\varepsilon},
    \label{eq:ncc}
\end{equation}
where $\tilde I = I-\mu_I$ and $\tilde J = J-\mu_J$ are the mean-centered volumes, $\mu_I$
and $\mu_J$ are their mean intensities, $\varepsilon=10^{-5}$ guards against division
by zero and every sum runs over all voxels $p\in\Omega$. 

The smoothness penalty is the
squared-$L_2$ first-order gradient of the predicted field, averaged over the
three spatial axes, the three field components and all voxels,
\begin{equation}
    \mathcal{L}_{\mathrm{smooth}}(u)
    \;=\;
    \frac{1}{9|\Omega|}
    \sum_{p\in\Omega}\sum_{d}\sum_{c}
    \bigl(\partial_d u_c(p)\bigr)^2,
    \label{eq:grad}
\end{equation}
discretized with first-order forward differences. Here $|\Omega|$ is the number of voxels
in the image domain, $u_c$ is the component of the field along axis $c$ and $\partial_d$
is the partial derivative along axis $d$, with both $c$ and $d$ running over the three
spatial axes $z$, $y$ and $x$. The penalty is therefore the squared first derivative
averaged over every axis, every field component and every voxel. It always acts on the
raw field $u$ emitted by the network.
Algorithm~\ref{alg:dlbr} summarizes the complete forward pass and training
step.

\begin{algorithm}[t]
\caption{DLBR forward pass and training step}\label{alg:dlbr}
\begin{algorithmic}[1]
\Require moving volume $I_m$, template $I_f$, network $g_\theta$,
smoothness weight $\lambda_{\mathrm{smooth}}$
\Ensure moved volume $I_m\circ\phi$, deformation $\phi$, loss $\mathcal{L}$
\State $x \gets \mathrm{concat}(I_m, I_f)$
\State $f \gets \mathrm{UNet}_\theta(x)$
\State $u \gets \mathrm{Conv}_{3\times3\times3}(f)$
\State $\phi \gets \mathrm{Id} + u$
\State $I_m\circ\phi \gets \mathrm{SpatialTransformer}(I_m, \phi)$
\State $\mathcal{L} \gets \mathcal{L}_{\mathrm{sim}}(I_m\circ\phi, I_f)
        + \lambda_{\mathrm{smooth}}\,\mathcal{L}_{\mathrm{smooth}}(u)$
\State update $\theta$ with Adam on $\nabla_\theta\mathcal{L}$
\end{algorithmic}
\end{algorithm}

\subsubsection{Comparable Deep Learning Methods}
We compare DLBR against seven further learned registration networks that span the dominant architectural families in the recent literature. Each one follows its reference implementation, adapted to our framework so that every method shares the same training and evaluation setup and differs only in the network $g_\theta$ that maps the stacked moving-and-fixed pair to a deformation field. TransMorph from \cite{chen2022transmorph} replaces the U-Net encoder with a Swin Transformer whose shifted-window self-attention captures long-range dependencies, while ViT-V-Net from \cite{chen2021vitvnet} tokenizes the deepest features of a convolutional stem and reasons over them with a Vision Transformer encoder. LH-Morph from \cite{sadegheih2024lhunet} uses the light hybrid LHU-Net backbone, a convolutional encoder-decoder in which the type of attention block is selected according to the depth level of the network, and MambaMorph from \cite{guo2024mambamorph} instead inserts selective state-space (Mamba) blocks for global context with linear memory cost. LapIRN from \cite{mok2020lapirn} and RDP from \cite{wang2024rdp} both adopt a coarse-to-fine pyramid that accumulates residual displacements across levels with a weight-shared network, capturing large shifts and fine deformation together. Fourier-Net from \cite{jia2023fouriernet} predicts the deformation in the frequency domain by retaining only learnable low-frequency modes, so band-limiting acts as an intrinsic smoothness prior.

\subsubsection{Implementation Details}

Each step pairs one Janelia volume with the shared template at a batch size
of one. The network is implemented in PyTorch and optimized with the
Adam optimizer of \cite{kingma2015adam} at a learning rate of $3\times10^{-4}$
under \texttt{bf16} mixed precision on a single NVIDIA
H100 GPU. We hold out a fixed, quality-stratified $10\%$ of the Janelia
volumes as a validation set, train for up to $500$ epochs, and stop early
once the validation loss has not improved for $20$ epochs, retaining the
checkpoint with the lowest validation loss. The Janelia train and
validation partition is held fixed across all runs so that every model is
exposed to the same split. The similarity loss term is the normalized
cross-correlation of Eq.~\ref{eq:ncc} and the smoothness weight is
$\lambda_{\mathrm{smooth}}=0.01$. The field predicted by the network is used directly as
the displacement of Eq.~\ref{eq:deformation}. These settings were fixed on the
Janelia validation split and applied unchanged to the held-out Larvalign test
set.

Algorithm~\ref{alg:pipeline} states what registering one new acquisition involves once
the network is trained. It covers the whole path from the raw microscope stack to the
moved volume, so the preprocessing of Section~\ref{subsec:preprocessing} and the network
forward pass appear as the single procedure a user actually runs. No optimization happens
anywhere in it beyond the fast coarse affine pre-alignment.

\begin{algorithm}[t]
\caption{DLBR registration of a new acquisition}\label{alg:pipeline}
\begin{algorithmic}[1]
\Require raw acquisition, template $I_f$, reference grid, trained network $g_\theta$
\Ensure moved volume $I_m\circ\phi$, deformation $\phi$
\State select the anatomical signal channel and transpose to $(Z,Y,X)$
\State attach the physical voxel spacing of the acquisition
\State clip to the $0.01$ and $99.9$ foreground percentiles, rescale to $[0,1]$
\State $A \gets \mathrm{AffineRegister}(\text{volume}, I_f)$
\State $I_m \gets \mathrm{Resample}(\text{volume}, A, \text{reference grid})$
\State $u \gets \mathrm{Conv}_{3\times3\times3}\bigl(\mathrm{UNet}_\theta(
        \mathrm{concat}(I_m, I_f))\bigr)$
\State $\phi \gets \mathrm{Id} + u$
\State $I_m\circ\phi \gets \mathrm{SpatialTransformer}(I_m, \phi)$
\State \Return $I_m\circ\phi$, $\phi$
\end{algorithmic}
\end{algorithm}

\subsection{Evaluation Metrics}\label{subsec:metrics}

The learned models and the classical baselines are scored under one identical
metric protocol, which we discuss in the following. All scores compare the moved volume
$I_m\circ\phi$ against the fixed template $I_f$. Two global scores measure intensity agreement over the whole volume. The
first is a normalized mutual information (NMI), taken in the overlap-invariant sense
introduced by \cite{studholme1999nmi} and used here in the symmetric form bounded in
$[0,1]$,
\begin{equation}
    \mathrm{NMI}\bigl(I_m\!\circ\!\phi,\,I_f\bigr)
    \;=\;
    \frac{2\sum_{a,b} p(a,b)\,\log\frac{p(a,b)}{p_X(a)\,p_Y(b)}}
         {H(I_m\circ\phi)+H(I_f)},
    \label{eq:nmi-global}
\end{equation}
where $p(a,b)$ is the joint intensity histogram, $p_X$
and $p_Y$ are its marginals and $H(\cdot)$ are the marginal entropies. We use normalization so that identical volumes score $1$.

The second global score is the
global Pearson cross-correlation in $[-1,1]$,
\begin{equation}
    \mathrm{NCC}\bigl(I_m\!\circ\!\phi,\,I_f\bigr)
    =
    \frac{\sum_{p} \tilde{I}_m(p)\,\tilde{I}_f(p)}
      {\sqrt{\sum_{p}\tilde{I}_m(p)^{2}}
       \sqrt{\sum_{p}\tilde{I}_f(p)^{2}}},
    \label{eq:ncc-global}
\end{equation}
where $\tilde{I}_m=(I_m\circ\phi)-\mu_m$ and $\tilde{I}_f=I_f-\mu_f$ are the
mean-centered volumes, $\mu_m$ and $\mu_f$ are their mean intensities and every
sum runs over all voxels $p\in\Omega$. For both global scores, higher is
better.

\begin{figure}[t]
\centering
\includegraphics[width=0.48\columnwidth]{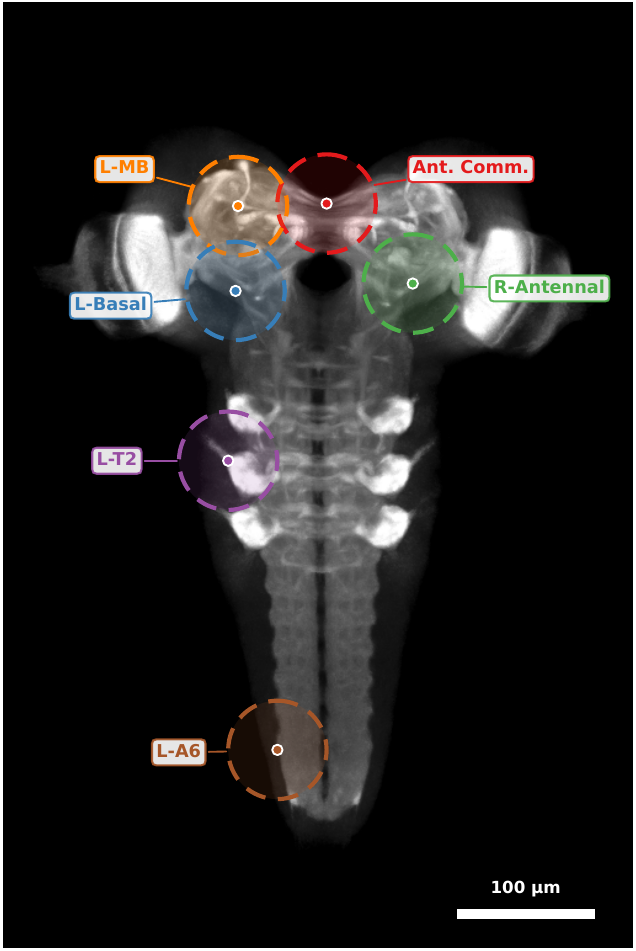}
\caption{\textbf{Landmark-local measurement regions on the template.} The landmark-local metrics are evaluated inside the
ellipsoidal neighborhoods $E_\ell$ of Eq.~\ref{eq:landmark-ellipsoid}. Six of the $L=30$
regions are drawn as examples.}
\label{fig:local-roi}
\end{figure}

To anchor accuracy to anatomy we use the $L=30$ landmarks of the reference template. Around each landmark
$c_\ell=(c_{\ell,z},c_{\ell,y},c_{\ell,x})$ we take an anisotropic
ellipsoidal neighborhood
\begin{multline}
    E_\ell \;=\; \Bigl\{p\in\Omega :
    \Bigl(\tfrac{p_z-c_{\ell,z}}{r_z}\Bigr)^{\!2}
    +\Bigl(\tfrac{p_y-c_{\ell,y}}{r_y}\Bigr)^{\!2} \\
    +\Bigl(\tfrac{p_x-c_{\ell,x}}{r_x}\Bigr)^{\!2}
    \leq 1\Bigr\},
    \label{eq:landmark-ellipsoid}
\end{multline}
with radii $(r_x,r_y,r_z)=(40,40,20)$ voxels held fixed across methods.
Figure~\ref{fig:local-roi} illustrates these measurement regions on
the template, with one representative ellipsoid drawn per anatomical group.
Inside $E_\ell$ we evaluate two complementary errors.

The first is
the local NMI of Eq.~\ref{eq:nmi-global} restricted to $E_\ell$, denoted $\mathrm{NMI}_{\mathrm{loc}}^{(\ell)}$. 

The second
is the local intensity mean squared error (MSE)
\begin{equation}
    \mathrm{MSE}_{\mathrm{loc}}^{(\ell)}
    \;=\;
    \frac{1}{|E_\ell|}\sum_{p\in E_\ell}
    \bigl(I_f(p) - (I_m\circ\phi)(p)\bigr)^2 .
    \label{eq:mse-local}
\end{equation}
Each is summarized per case as an arithmetic mean and a voxel-count-weighted
mean over the valid landmarks, for example
\begin{equation}
    \overline{\mathrm{NMI}}_{\mathrm{loc}}
    \;=\;
    \frac{1}{L}\sum_{\ell=1}^{L}
    \mathrm{NMI}_{\mathrm{loc}}^{(\ell)},
    \label{eq:nmi-local-mean}
\end{equation}
and is also retained for every individual landmark. 

The mean local NMI is the primary accuracy axis in the results. Larval brain volumes are dominated by dark background, so a whole-volume
score is driven largely by voxels that carry no anatomy at all. Restricting the comparison
to the expert-annotated landmark neighborhoods measures agreement exactly where the
anatomy is, and mutual information tolerates the staining-intensity differences between
specimens that a direct intensity difference would penalize.

\subsection{Statistical Analysis}\label{subsec:stats}
All statistics are computed over per-case results, where one case is one test volume scored
under the protocol of Section~\ref{subsec:metrics}. Aggregates are formed per method, per
resolution scale and per acquisition-quality stratum, as well as over the complete test
set. Each aggregate is the mean over its cases together with the standard deviation across
them, so the reported spread describes how much a metric varies from one specimen to the
next. Confidence intervals come from a non-parametric bootstrap over the per-case results,
drawing $2000$ resamples with replacement and taking the $95\%$ percentile interval between
the $2.5$ and $97.5$ points.

Significance is assessed at $\alpha=0.05$ with two-sided paired Wilcoxon signed-rank tests
on the per-case mean local NMI. Cases are matched by index, so every comparison is paired
on the same volumes, and the tests are run within each acquisition-quality stratum as well
as on the complete test set. Two comparisons are drawn per stratum. The first places DLBR
against the runner-up learned method in that stratum, and the second places the weakest
learned method against the strongest classical method in that stratum, so the two tests
together cover the margin at the top of the ranking and the margin at the boundary between
the learned and the classical family. Bonferroni correction is applied over the full set of
comparisons drawn.

The same bootstrap supports two further analyses. The fraction of accuracy a method loses
as acquisition quality degrades is given a confidence interval by resampling each quality
stratum independently. The configuration ablation varies one training setting at a time and
quantifies each setting by the change in mean local NMI relative to the featured
configuration, with a bootstrap interval on that change, so the magnitude of an effect is
visible alongside its uncertainty.

\section{Results}\label{sec:results}

We evaluate every method under the single shared protocol of
Section~\ref{subsec:metrics}. All methods register the same preprocessed and
template pre-aligned volumes at the highest preprocessing resolution and are scored on
the same held-out Larvalign test set. Because DLBR is trained only on Janelia volumes and
tested only on the Larvalign collection, every accuracy number reported below is an
out-of-distribution result. The section is organized as follows.
Section~\ref{subsec:accuracy} measures overall registration accuracy across all methods.
Section~\ref{subsec:robustness} asks how much of that accuracy survives when the
acquisition quality degrades. Section~\ref{subsec:per-landmark} resolves the result into
the individual anatomical structures, and Section~\ref{subsec:speed} turns from accuracy
to the cost of registering one volume. Section~\ref{subsec:qualitative} inspects the
alignments themselves and provides qualitative results. Section~\ref{subsec:ablation} reports two ablations, one over
the DLBR training configuration and one over the preprocessing resolution at which the
main results are reported. Section~\ref{subsec:limitations} closes with the memory bound
that sets the highest resolution at which the model can be trained.

\begin{figure*}[tp]
\centering
\includegraphics[width=0.8\textwidth,height=0.9\textheight,keepaspectratio]{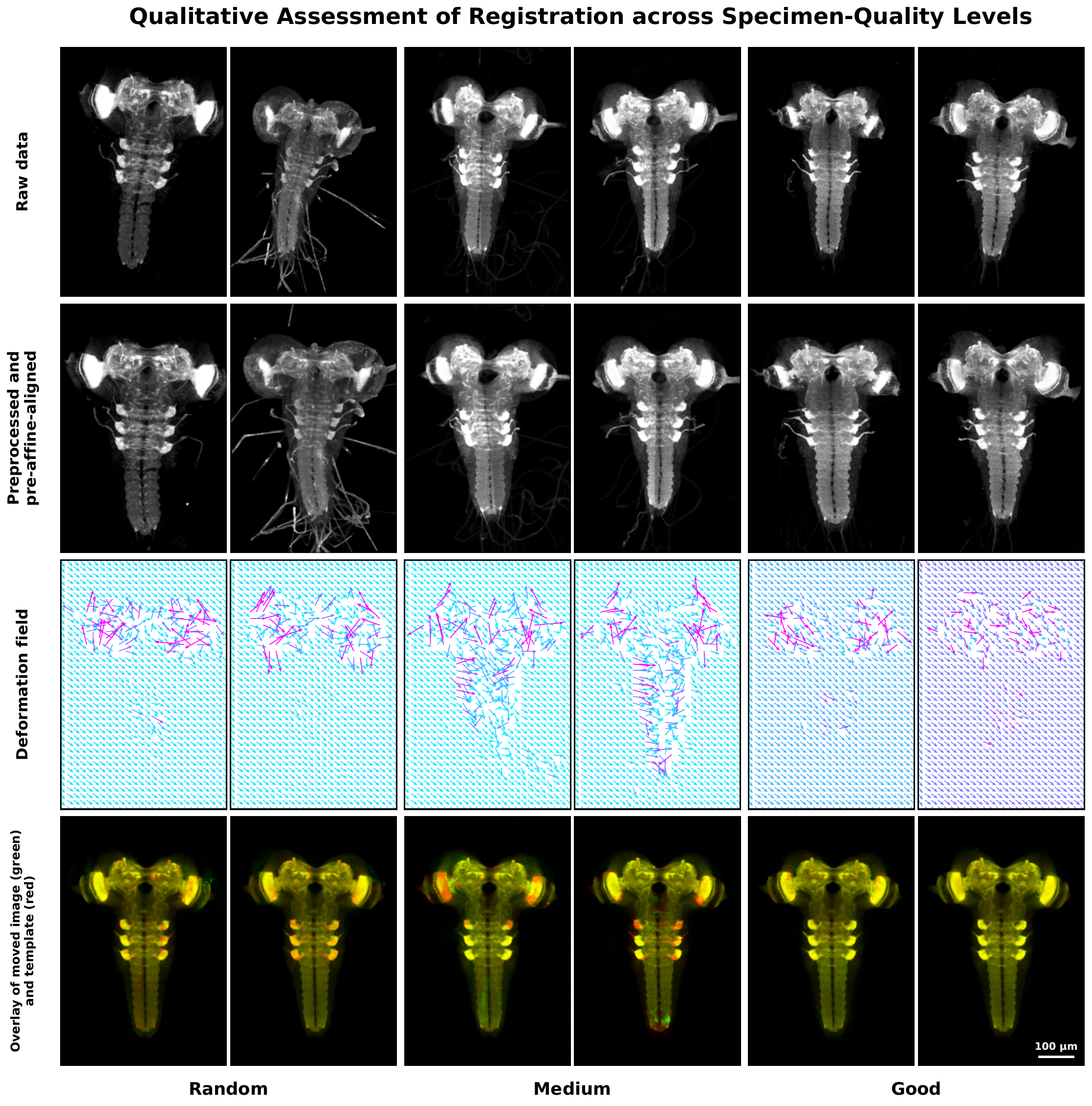}
\caption{\textbf{Qualitative DLBR registration on six representative test cases.}
Maximum-resolution results, columns ordered from random to good quality. Rows show the raw
resampled input, the preprocessed and affine pre-aligned input, the predicted deformation
field, and the overlay of the moved volume in green on the template in red, where yellow
marks agreement. All panels are $Z$ maximum-intensity projections.}
\label{fig:qualitative_overlay}
\end{figure*}

\subsection{Registration accuracy across methods}\label{subsec:accuracy}

Table~\ref{tab:metric_global_nmi} and Table~\ref{tab:metric_local_nmi} report the
global (cf. Eq.~\ref{eq:nmi-global}) and landmark-local
(cf. Eq.~\ref{eq:nmi-local-mean}) normalized mutual information,
Table~\ref{tab:metric_local_mse} the landmark-local intensity mean squared error
(cf. Eq.~\ref{eq:mse-local}), and Table~\ref{tab:metric_global_ncc} the global
normalized cross-correlation (cf. Eq.~\ref{eq:ncc-global}). The two
landmark-local metrics are both evaluated inside the neighborhoods of
Eq.~\ref{eq:landmark-ellipsoid}. Three findings hold across all four metrics.

First, DLBR provides the most accurate alignment (i.e. the highest landmark-local NMI) among the evaluated methods. It attains a complete-set landmark-local NMI of $0.508$, a global NMI of $0.585$ and a
global NCC of $0.958$, the best value in each case. Its margin over the strongest
classical method is $0.230$ of local NMI,
a gain of roughly $23$ percentage points over the $0.278$ reached by diffeomorphic
Demons.

Second, the gain is one of learned over classical registration rather than of one
particular network. On the complete test set every learned network exceeds the
strongest classical baseline on both the global and the landmark-local intensity
agreement. Among the classical methods the deformable ones lead, with diffeomorphic
Demons reaching a local NMI of $0.278$ and NiftyReg F3D and ANTs SyN tied behind it at
$0.245$, while the affine and rigid baselines stay close to the
no-registration identity floor of $0.144$. All learned networks except the two
pyramidal ones clear the best classical value by a wide margin, so the separation
between the two families is far larger than the spread within the learned family.

Third, the learned methods show generally higher performance, with DLBR and MambaMorph forming the top two methods. MambaMorph is the closest competitor throughout, trailing DLBR by $0.036$ in complete-set landmark-local NMI and
by $0.028$ in global NMI, and it takes the second position in almost every comparison. The
remaining transformer and hybrid networks, namely ViT-V-Net, LH-Morph, TransMorph and
Fourier-Net, form a consistent second tier, and the pyramidal LapIRN and RDP trail the
other learned methods on intensity agreement while still improving on every classical
baseline. The closest the field comes to DLBR is the global NCC within the
medium-quality stratum, where DLBR and ViT-V-Net are on par at $0.942$ against
$0.943$.

\begin{table*}[!htbp]
\centering
\footnotesize
\caption{\textbf{Whole-volume mutual information on the external Larvalign test set.} Per-case whole-volume NMI (cf. Eq.~\ref{eq:nmi-global}), stratified by image quality and over the complete set. Cells report mean $\pm$ standard deviation over test cases, higher is better. Best value per column in \textbf{bold}, second best \underline{underlined}.}
\label{tab:metric_global_nmi}
\begin{tabular}{c l rrrr}
\toprule[1.2pt]
& & \multicolumn{4}{c}{\textbf{Global NMI} $\uparrow$} \\
\cmidrule(lr){3-6}
& \textbf{Method} & \multicolumn{1}{c}{Random} & \multicolumn{1}{c}{Medium} & \multicolumn{1}{c}{Good} & \multicolumn{1}{c}{All} \\
\midrule
\multirow{11}{*}{\rotatebox[origin=c]{90}{\textbf{\shortstack{Classical\\Methods}}}} & Identity & 0.226 $\pm$ 0.069 & 0.221 $\pm$ 0.084 & 0.311 $\pm$ 0.090 & 0.250 $\pm$ 0.090 \\
 & Rigid (SimpleITK)~\yrcite{lowekamp2013simpleitk} & 0.233 $\pm$ 0.067 & 0.226 $\pm$ 0.083 & 0.314 $\pm$ 0.091 & 0.256 $\pm$ 0.089 \\
 & Affine (SimpleITK)~\yrcite{lowekamp2013simpleitk} & 0.228 $\pm$ 0.064 & 0.222 $\pm$ 0.074 & 0.324 $\pm$ 0.099 & 0.255 $\pm$ 0.091 \\
 & B-spline (SimpleITK)~\yrcite{Rueckert1999} & 0.309 $\pm$ 0.071 & 0.337 $\pm$ 0.039 & 0.388 $\pm$ 0.062 & 0.342 $\pm$ 0.068 \\
 & Demons (SimpleITK)~\yrcite{vercauteren2009demons} & 0.346 $\pm$ 0.083 & 0.335 $\pm$ 0.103 & 0.447 $\pm$ 0.103 & 0.373 $\pm$ 0.108 \\
 & Elastix rigid~\yrcite{klein2010elastix} & 0.236 $\pm$ 0.067 & 0.240 $\pm$ 0.074 & 0.316 $\pm$ 0.089 & 0.261 $\pm$ 0.085 \\
 & Elastix affine~\yrcite{klein2010elastix} & 0.234 $\pm$ 0.064 & 0.232 $\pm$ 0.078 & 0.324 $\pm$ 0.093 & 0.261 $\pm$ 0.089 \\
 & Elastix B-spline~\yrcite{klein2010elastix} & 0.257 $\pm$ 0.073 & 0.285 $\pm$ 0.051 & 0.353 $\pm$ 0.084 & 0.295 $\pm$ 0.081 \\
 & Elastix affine+B-spline~\yrcite{klein2010elastix} & 0.256 $\pm$ 0.071 & 0.278 $\pm$ 0.056 & 0.354 $\pm$ 0.083 & 0.293 $\pm$ 0.082 \\
 & ANTs SyN~\yrcite{avants2008syn} & 0.298 $\pm$ 0.087 & 0.287 $\pm$ 0.116 & 0.398 $\pm$ 0.115 & 0.325 $\pm$ 0.116 \\
 & NiftyReg F3D~\yrcite{modat2010niftyreg} & 0.350 $\pm$ 0.077 & 0.359 $\pm$ 0.060 & 0.423 $\pm$ 0.089 & 0.375 $\pm$ 0.082 \\
\midrule
\multirow{8}{*}{\rotatebox[origin=c]{90}{\textbf{\shortstack{Deep-learning\\Methods}}}} & TransMorph~\yrcite{chen2022transmorph} & 0.488 $\pm$ 0.078 & 0.476 $\pm$ 0.059 & 0.540 $\pm$ 0.075 & 0.500 $\pm$ 0.076 \\
 & LapIRN~\yrcite{mok2020lapirn} & 0.371 $\pm$ 0.082 & 0.340 $\pm$ 0.123 & 0.460 $\pm$ 0.116 & 0.388 $\pm$ 0.117 \\
 & RDP~\yrcite{wang2024rdp} & 0.358 $\pm$ 0.081 & 0.331 $\pm$ 0.118 & 0.449 $\pm$ 0.114 & 0.377 $\pm$ 0.115 \\
 & Fourier-Net~\yrcite{jia2023fouriernet} & 0.485 $\pm$ 0.099 & 0.452 $\pm$ 0.101 & 0.546 $\pm$ 0.108 & 0.493 $\pm$ 0.109 \\
 & ViT-V-Net~\yrcite{chen2021vitvnet} & 0.534 $\pm$ 0.076 & \underline{0.529} $\pm$ 0.046 & 0.580 $\pm$ 0.068 & 0.546 $\pm$ 0.069 \\
 & MambaMorph~\yrcite{guo2024mambamorph} & \underline{0.549} $\pm$ 0.082 & \underline{0.529} $\pm$ 0.071 & \underline{0.595} $\pm$ 0.082 & \underline{0.557} $\pm$ 0.083 \\
 & LH-Morph~\yrcite{sadegheih2024lhunet} & 0.498 $\pm$ 0.090 & 0.488 $\pm$ 0.063 & 0.552 $\pm$ 0.086 & 0.511 $\pm$ 0.086 \\
\cmidrule(lr){2-6}
 & \textbf{DLBR} & \textbf{0.579} $\pm$ 0.081 & \textbf{0.554} $\pm$ 0.079 & \textbf{0.626} $\pm$ 0.085 & \textbf{0.585} $\pm$ 0.086 \\
\bottomrule[1.2pt]
\end{tabular}
\end{table*}

\begin{table*}[!htbp]
\centering
\footnotesize
\caption{\textbf{Landmark-local mutual information on the external Larvalign test set.} Per-case mean landmark-local NMI (cf. Eq.~\ref{eq:nmi-local-mean}) across the small local neighborhoods of the 30 anatomical landmarks (cf. Eq.~\ref{eq:landmark-ellipsoid}), stratified by image quality and over the complete set. Cells report mean $\pm$ standard deviation over test cases, higher is better. Best value per column in \textbf{bold}, second best \underline{underlined}.}
\label{tab:metric_local_nmi}
\begin{tabular}{c l rrrr}
\toprule[1.2pt]
& & \multicolumn{4}{c}{\textbf{Local NMI} $\uparrow$} \\
\cmidrule(lr){3-6}
& \textbf{Method} & \multicolumn{1}{c}{Random} & \multicolumn{1}{c}{Medium} & \multicolumn{1}{c}{Good} & \multicolumn{1}{c}{All} \\
\midrule
\multirow{11}{*}{\rotatebox[origin=c]{90}{\textbf{\shortstack{Classical\\Methods}}}} & Identity & 0.122 $\pm$ 0.056 & 0.127 $\pm$ 0.049 & 0.191 $\pm$ 0.063 & 0.144 $\pm$ 0.064 \\
 & Rigid (SimpleITK)~\yrcite{lowekamp2013simpleitk} & 0.128 $\pm$ 0.053 & 0.132 $\pm$ 0.047 & 0.193 $\pm$ 0.063 & 0.149 $\pm$ 0.062 \\
 & Affine (SimpleITK)~\yrcite{lowekamp2013simpleitk} & 0.120 $\pm$ 0.053 & 0.128 $\pm$ 0.042 & 0.201 $\pm$ 0.072 & 0.147 $\pm$ 0.067 \\
 & B-spline (SimpleITK)~\yrcite{Rueckert1999} & 0.170 $\pm$ 0.081 & 0.204 $\pm$ 0.034 & 0.262 $\pm$ 0.072 & 0.209 $\pm$ 0.077 \\
 & Demons (SimpleITK)~\yrcite{vercauteren2009demons} & 0.247 $\pm$ 0.091 & 0.244 $\pm$ 0.089 & 0.353 $\pm$ 0.091 & 0.278 $\pm$ 0.103 \\
 & Elastix rigid~\yrcite{klein2010elastix} & 0.129 $\pm$ 0.053 & 0.140 $\pm$ 0.042 & 0.196 $\pm$ 0.064 & 0.153 $\pm$ 0.061 \\
 & Elastix affine~\yrcite{klein2010elastix} & 0.118 $\pm$ 0.049 & 0.133 $\pm$ 0.043 & 0.201 $\pm$ 0.068 & 0.148 $\pm$ 0.065 \\
 & Elastix B-spline~\yrcite{klein2010elastix} & 0.141 $\pm$ 0.062 & 0.155 $\pm$ 0.044 & 0.236 $\pm$ 0.080 & 0.174 $\pm$ 0.076 \\
 & Elastix affine+B-spline~\yrcite{klein2010elastix} & 0.137 $\pm$ 0.063 & 0.158 $\pm$ 0.042 & 0.237 $\pm$ 0.079 & 0.174 $\pm$ 0.076 \\
 & ANTs SyN~\yrcite{avants2008syn} & 0.210 $\pm$ 0.095 & 0.214 $\pm$ 0.090 & 0.322 $\pm$ 0.100 & 0.245 $\pm$ 0.108 \\
 & NiftyReg F3D~\yrcite{modat2010niftyreg} & 0.212 $\pm$ 0.092 & 0.222 $\pm$ 0.064 & 0.312 $\pm$ 0.096 & 0.245 $\pm$ 0.096 \\
\midrule
\multirow{8}{*}{\rotatebox[origin=c]{90}{\textbf{\shortstack{Deep-learning\\Methods}}}} & TransMorph~\yrcite{chen2022transmorph} & 0.381 $\pm$ 0.078 & 0.366 $\pm$ 0.079 & 0.444 $\pm$ 0.065 & 0.395 $\pm$ 0.081 \\
 & LapIRN~\yrcite{mok2020lapirn} & 0.278 $\pm$ 0.090 & 0.259 $\pm$ 0.098 & 0.373 $\pm$ 0.095 & 0.301 $\pm$ 0.106 \\
 & RDP~\yrcite{wang2024rdp} & 0.265 $\pm$ 0.088 & 0.250 $\pm$ 0.095 & 0.361 $\pm$ 0.093 & 0.289 $\pm$ 0.103 \\
 & Fourier-Net~\yrcite{jia2023fouriernet} & 0.397 $\pm$ 0.097 & 0.360 $\pm$ 0.112 & 0.473 $\pm$ 0.089 & 0.408 $\pm$ 0.110 \\
 & ViT-V-Net~\yrcite{chen2021vitvnet} & 0.431 $\pm$ 0.078 & 0.425 $\pm$ 0.071 & 0.492 $\pm$ 0.059 & 0.448 $\pm$ 0.076 \\
 & MambaMorph~\yrcite{guo2024mambamorph} & \underline{0.461} $\pm$ 0.084 & \underline{0.437} $\pm$ 0.090 & \underline{0.522} $\pm$ 0.070 & \underline{0.472} $\pm$ 0.089 \\
 & LH-Morph~\yrcite{sadegheih2024lhunet} & 0.394 $\pm$ 0.093 & 0.389 $\pm$ 0.076 & 0.468 $\pm$ 0.067 & 0.415 $\pm$ 0.088 \\
\cmidrule(lr){2-6}
 & \textbf{DLBR} & \textbf{0.498} $\pm$ 0.085 & \textbf{0.472} $\pm$ 0.093 & \textbf{0.559} $\pm$ 0.071 & \textbf{0.508} $\pm$ 0.091 \\
\bottomrule[1.2pt]
\end{tabular}
\end{table*}

\begin{table*}[!htbp]
\centering
\footnotesize
\caption{\textbf{Landmark-local intensity error on the external Larvalign test set.} Per-case mean landmark-local intensity MSE (cf. Eq.~\ref{eq:mse-local}) across the small local neighborhoods of the 30 anatomical landmarks (cf. Eq.~\ref{eq:landmark-ellipsoid}), stratified by image quality and over the complete set. Cells report mean $\pm$ standard deviation over test cases, lower is better. Best value per column in \textbf{bold}, second best \underline{underlined}.}
\label{tab:metric_local_mse}
\begin{tabular}{c l rrrr}
\toprule[1.2pt]
& & \multicolumn{4}{c}{\textbf{Local MSE} $\downarrow$} \\
\cmidrule(lr){3-6}
& \textbf{Method} & \multicolumn{1}{c}{Random} & \multicolumn{1}{c}{Medium} & \multicolumn{1}{c}{Good} & \multicolumn{1}{c}{All} \\
\midrule
\multirow{11}{*}{\rotatebox[origin=c]{90}{\textbf{\shortstack{Classical\\Methods}}}} & Identity & 0.0223 $\pm$ 0.0086 & 0.0281 $\pm$ 0.0092 & 0.0210 $\pm$ 0.0086 & 0.0237 $\pm$ 0.0093 \\
 & Rigid (SimpleITK)~\yrcite{lowekamp2013simpleitk} & 0.0217 $\pm$ 0.0083 & 0.0279 $\pm$ 0.0091 & 0.0207 $\pm$ 0.0085 & 0.0234 $\pm$ 0.0092 \\
 & Affine (SimpleITK)~\yrcite{lowekamp2013simpleitk} & 0.0226 $\pm$ 0.0085 & 0.0304 $\pm$ 0.0093 & 0.0205 $\pm$ 0.0101 & 0.0245 $\pm$ 0.0101 \\
 & B-spline (SimpleITK)~\yrcite{Rueckert1999} & 0.0263 $\pm$ 0.0171 & 0.0339 $\pm$ 0.0123 & 0.0201 $\pm$ 0.0138 & 0.0269 $\pm$ 0.0157 \\
 & Demons (SimpleITK)~\yrcite{vercauteren2009demons} & 0.0110 $\pm$ 0.0051 & 0.0137 $\pm$ 0.0047 & 0.0080 $\pm$ 0.0047 & 0.0109 $\pm$ 0.0053 \\
 & Elastix rigid~\yrcite{klein2010elastix} & 0.0216 $\pm$ 0.0082 & 0.0287 $\pm$ 0.0092 & 0.0208 $\pm$ 0.0089 & 0.0236 $\pm$ 0.0094 \\
 & Elastix affine~\yrcite{klein2010elastix} & 0.0226 $\pm$ 0.0079 & 0.0315 $\pm$ 0.0112 & 0.0204 $\pm$ 0.0104 & 0.0248 $\pm$ 0.0109 \\
 & Elastix B-spline~\yrcite{klein2010elastix} & 0.0208 $\pm$ 0.0087 & 0.0324 $\pm$ 0.0124 & 0.0185 $\pm$ 0.0120 & 0.0238 $\pm$ 0.0125 \\
 & Elastix affine+B-spline~\yrcite{klein2010elastix} & 0.0211 $\pm$ 0.0085 & 0.0337 $\pm$ 0.0132 & 0.0187 $\pm$ 0.0127 & 0.0244 $\pm$ 0.0131 \\
 & ANTs SyN~\yrcite{avants2008syn} & 0.0150 $\pm$ 0.0088 & 0.0181 $\pm$ 0.0090 & 0.0110 $\pm$ 0.0081 & 0.0148 $\pm$ 0.0091 \\
 & NiftyReg F3D~\yrcite{modat2010niftyreg} & 0.0171 $\pm$ 0.0104 & 0.0225 $\pm$ 0.0123 & 0.0130 $\pm$ 0.0106 & 0.0175 $\pm$ 0.0117 \\
\midrule
\multirow{8}{*}{\rotatebox[origin=c]{90}{\textbf{\shortstack{Deep-learning\\Methods}}}} & TransMorph~\yrcite{chen2022transmorph} & 0.0036 $\pm$ 0.0018 & 0.0053 $\pm$ 0.0025 & 0.0034 $\pm$ 0.0019 & 0.0041 $\pm$ 0.0022 \\
 & LapIRN~\yrcite{mok2020lapirn} & 0.0100 $\pm$ 0.0053 & 0.0153 $\pm$ 0.0078 & 0.0090 $\pm$ 0.0062 & 0.0114 $\pm$ 0.0070 \\
 & RDP~\yrcite{wang2024rdp} & 0.0102 $\pm$ 0.0051 & 0.0147 $\pm$ 0.0068 & 0.0091 $\pm$ 0.0056 & 0.0113 $\pm$ 0.0063 \\
 & Fourier-Net~\yrcite{jia2023fouriernet} & 0.0044 $\pm$ 0.0028 & 0.0063 $\pm$ 0.0033 & 0.0032 $\pm$ 0.0024 & 0.0047 $\pm$ 0.0031 \\
 & ViT-V-Net~\yrcite{chen2021vitvnet} & 0.0028 $\pm$ 0.0016 & 0.0041 $\pm$ 0.0018 & 0.0027 $\pm$ 0.0015 & 0.0032 $\pm$ 0.0017 \\
 & MambaMorph~\yrcite{guo2024mambamorph} & \underline{0.0024} $\pm$ 0.0015 & \underline{0.0037} $\pm$ 0.0020 & \underline{0.0024} $\pm$ 0.0014 & \underline{0.0028} $\pm$ 0.0017 \\
 & LH-Morph~\yrcite{sadegheih2024lhunet} & 0.0033 $\pm$ 0.0019 & 0.0040 $\pm$ 0.0018 & \underline{0.0024} $\pm$ 0.0015 & 0.0032 $\pm$ 0.0019 \\
\cmidrule(lr){2-6}
 & \textbf{DLBR} & \textbf{0.0019} $\pm$ 0.0013 & \textbf{0.0026} $\pm$ 0.0016 & \textbf{0.0013} $\pm$ 0.0011 & \textbf{0.0020} $\pm$ 0.0015 \\
\bottomrule[1.2pt]
\end{tabular}
\end{table*}

\begin{table*}[!htbp]
\centering
\footnotesize
\caption{\textbf{Whole-volume cross-correlation on the external Larvalign test set.} Per-case whole-volume NCC (cf. Eq.~\ref{eq:ncc-global}), stratified by image quality and over the complete set. Cells report mean $\pm$ standard deviation over test cases, higher is better. Best value per column in \textbf{bold}, second best \underline{underlined}.}
\label{tab:metric_global_ncc}
\begin{tabular}{c l rrrr}
\toprule[1.2pt]
& & \multicolumn{4}{c}{\textbf{Global NCC} $\uparrow$} \\
\cmidrule(lr){3-6}
& \textbf{Method} & \multicolumn{1}{c}{Random} & \multicolumn{1}{c}{Medium} & \multicolumn{1}{c}{Good} & \multicolumn{1}{c}{All} \\
\midrule
\multirow{11}{*}{\rotatebox[origin=c]{90}{\textbf{\shortstack{Classical\\Methods}}}} & Identity & 0.552 $\pm$ 0.149 & 0.550 $\pm$ 0.129 & 0.704 $\pm$ 0.156 & 0.597 $\pm$ 0.161 \\
 & Rigid (SimpleITK)~\yrcite{lowekamp2013simpleitk} & 0.561 $\pm$ 0.146 & 0.559 $\pm$ 0.129 & 0.706 $\pm$ 0.155 & 0.604 $\pm$ 0.159 \\
 & Affine (SimpleITK)~\yrcite{lowekamp2013simpleitk} & 0.550 $\pm$ 0.147 & 0.547 $\pm$ 0.111 & 0.711 $\pm$ 0.166 & 0.598 $\pm$ 0.161 \\
 & B-spline (SimpleITK)~\yrcite{Rueckert1999} & 0.654 $\pm$ 0.143 & 0.701 $\pm$ 0.058 & 0.805 $\pm$ 0.120 & 0.715 $\pm$ 0.131 \\
 & Demons (SimpleITK)~\yrcite{vercauteren2009demons} & 0.760 $\pm$ 0.118 & 0.730 $\pm$ 0.134 & 0.861 $\pm$ 0.124 & 0.781 $\pm$ 0.137 \\
 & Elastix rigid~\yrcite{klein2010elastix} & 0.563 $\pm$ 0.144 & 0.582 $\pm$ 0.116 & 0.710 $\pm$ 0.153 & 0.613 $\pm$ 0.153 \\
 & Elastix affine~\yrcite{klein2010elastix} & 0.558 $\pm$ 0.148 & 0.575 $\pm$ 0.125 & 0.721 $\pm$ 0.155 & 0.613 $\pm$ 0.161 \\
 & Elastix B-spline~\yrcite{klein2010elastix} & 0.601 $\pm$ 0.161 & 0.654 $\pm$ 0.090 & 0.774 $\pm$ 0.150 & 0.670 $\pm$ 0.156 \\
 & Elastix affine+B-spline~\yrcite{klein2010elastix} & 0.594 $\pm$ 0.166 & 0.651 $\pm$ 0.093 & 0.776 $\pm$ 0.145 & 0.667 $\pm$ 0.159 \\
 & ANTs SyN~\yrcite{avants2008syn} & 0.696 $\pm$ 0.167 & 0.680 $\pm$ 0.172 & 0.830 $\pm$ 0.167 & 0.732 $\pm$ 0.181 \\
 & NiftyReg F3D~\yrcite{modat2010niftyreg} & 0.713 $\pm$ 0.155 & 0.773 $\pm$ 0.090 & 0.858 $\pm$ 0.129 & 0.776 $\pm$ 0.142 \\
\midrule
\multirow{8}{*}{\rotatebox[origin=c]{90}{\textbf{\shortstack{Deep-learning\\Methods}}}} & TransMorph~\yrcite{chen2022transmorph} & 0.932 $\pm$ 0.038 & 0.900 $\pm$ 0.053 & 0.944 $\pm$ 0.049 & 0.925 $\pm$ 0.050 \\
 & LapIRN~\yrcite{mok2020lapirn} & 0.795 $\pm$ 0.111 & 0.757 $\pm$ 0.129 & 0.877 $\pm$ 0.113 & 0.808 $\pm$ 0.127 \\
 & RDP~\yrcite{wang2024rdp} & 0.786 $\pm$ 0.114 & 0.749 $\pm$ 0.129 & 0.870 $\pm$ 0.115 & 0.800 $\pm$ 0.129 \\
 & Fourier-Net~\yrcite{jia2023fouriernet} & 0.912 $\pm$ 0.058 & 0.864 $\pm$ 0.079 & 0.928 $\pm$ 0.082 & 0.902 $\pm$ 0.077 \\
 & ViT-V-Net~\yrcite{chen2021vitvnet} & 0.952 $\pm$ 0.028 & \textbf{0.943} $\pm$ 0.024 & \underline{0.964} $\pm$ 0.027 & \underline{0.953} $\pm$ 0.028 \\
 & MambaMorph~\yrcite{guo2024mambamorph} & \underline{0.955} $\pm$ 0.027 & 0.935 $\pm$ 0.037 & \underline{0.964} $\pm$ 0.035 & 0.951 $\pm$ 0.035 \\
 & LH-Morph~\yrcite{sadegheih2024lhunet} & 0.931 $\pm$ 0.043 & 0.914 $\pm$ 0.044 & 0.948 $\pm$ 0.045 & 0.931 $\pm$ 0.046 \\
\cmidrule(lr){2-6}
 & \textbf{DLBR} & \textbf{0.961} $\pm$ 0.027 & \underline{0.942} $\pm$ 0.032 & \textbf{0.969} $\pm$ 0.031 & \textbf{0.958} $\pm$ 0.032 \\
\bottomrule[1.2pt]
\end{tabular}
\end{table*}

\subsection{Statistical significance and robustness to image quality}\label{subsec:robustness}

\begin{figure*}[t]
\centering
\includegraphics[width=\textwidth]{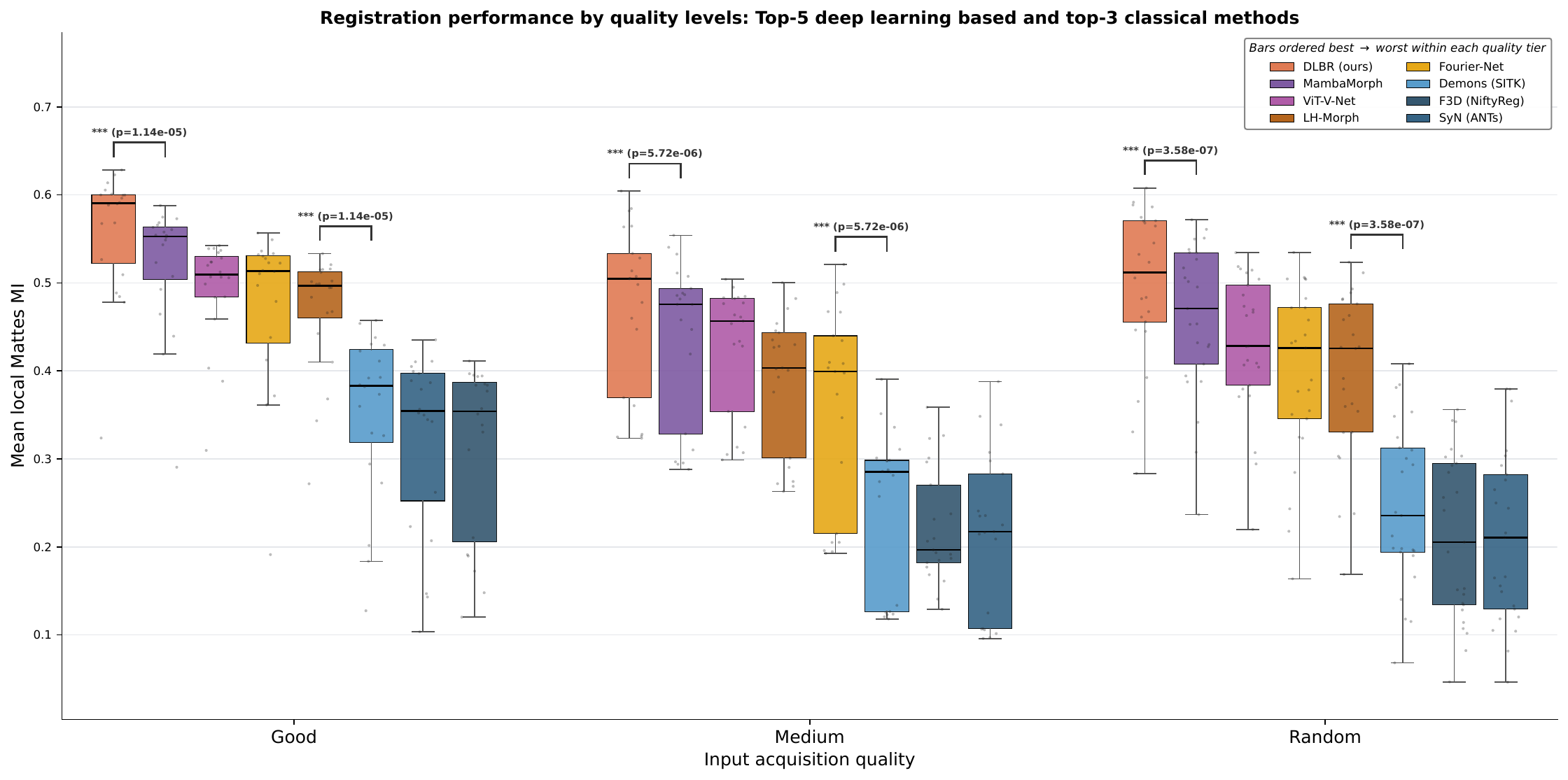}
\caption{\textbf{Registration accuracy by input acquisition quality tier.} Per-case
landmark-local NMI over the 30 landmarks on the 66 Larvalign test cases
(cf. Eq.~\ref{eq:nmi-local-mean}), grouped into good, medium and random tiers. Each tier shows the five strongest
learned methods, DLBR leftmost, then the three strongest classical methods, as
per-case boxplots with jittered points. Brackets are paired Wilcoxon signed-rank tests,
Bonferroni-corrected. Higher is better.}
\label{fig:quality_boxplot_stats}
\end{figure*}

Figure~\ref{fig:quality_boxplot_stats} resolves the accuracy comparison within each
acquisition-quality tier and attaches the significance tests defined in
Section~\ref{subsec:stats}. Within every tier we run paired Wilcoxon signed-rank
tests with Bonferroni correction across the family of comparisons. DLBR substantically outperforms the strongest classical baseline in the good, medium and random strata alike,
and the same holds for the learned methods as a group, so the gain over classical
registration is not confined to the easiest acquisitions. The per-case boxes also show
that the learned methods carry a much tighter case-to-case spread than the classical
deformable methods, so the advantage is not driven by a small number of easy cases.

\begin{figure*}[tp]
\centering
\includegraphics[width=0.8\textwidth,keepaspectratio]{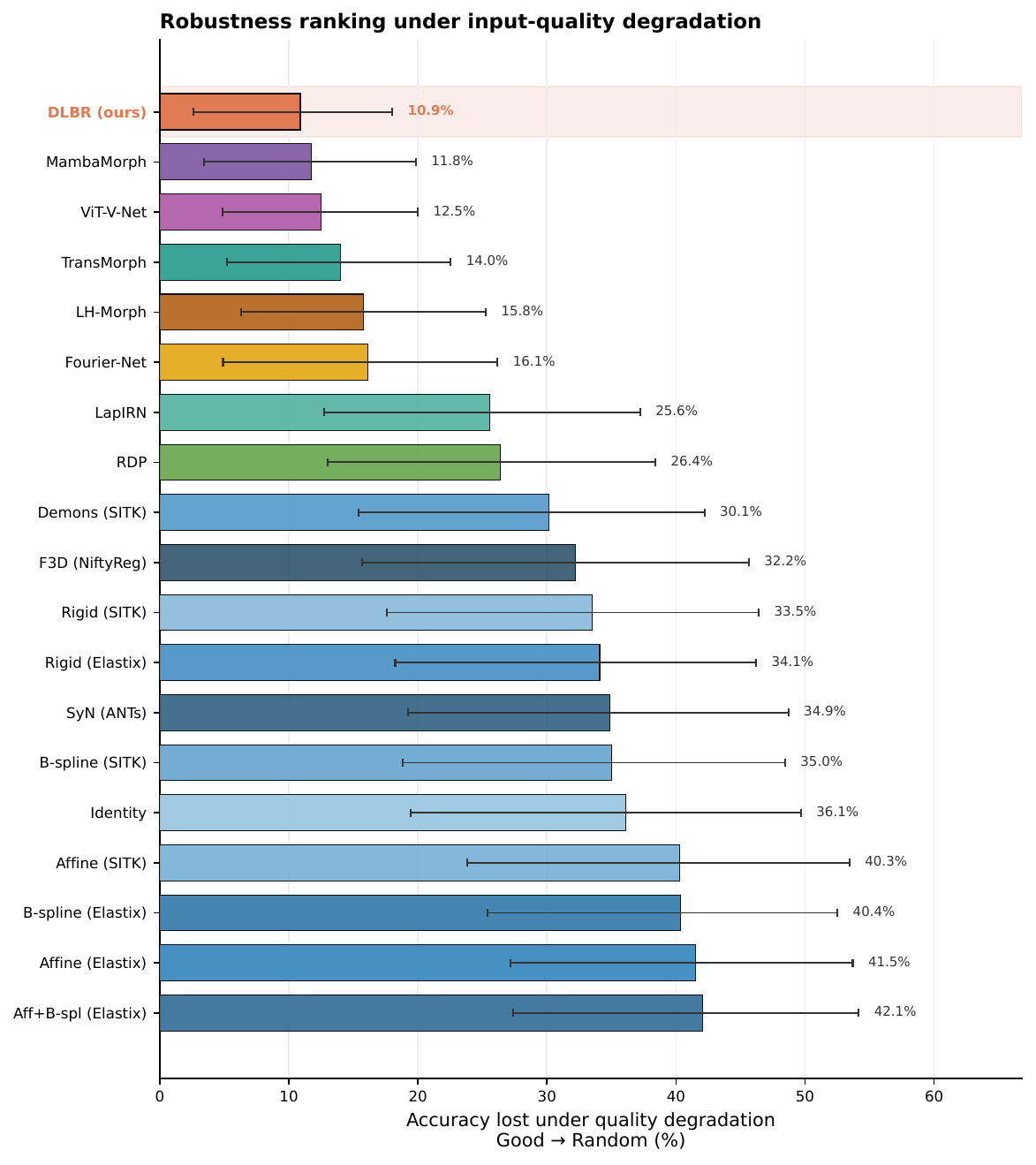}
\caption{\textbf{Robustness of accuracy to input acquisition quality.} Each bar is the
percentage of landmark-local NMI (cf. Eq.~\ref{eq:nmi-local-mean}) a method loses from
good-quality to random-quality inputs, sorted from most to least robust with DLBR
highlighted. Error bars are bootstrap 95 percent confidence intervals. Lower loss is
better.}
\label{fig:robustness_ranking_raw}
\end{figure*}

Figure~\ref{fig:robustness_ranking_raw} makes the robustness to input quality explicit.
Each bar is the fraction of landmark-local NMI a method loses when the input degrades
from good to random quality. DLBR is the most robust method on this axis, losing
$10.9$ percent. The five learned methods behind it lose between $12$ and $16$ percent
and the two pyramidal networks between $25$ and $27$ percent, whereas the classical
deformable methods lose between $30$ and $42$ percent. Since the entire Larvalign collection is external to the training distribution, this smaller degradation suggests that a network trained on one laboratory's data can transfer to volumes acquired under different imaging conditions and retain its performance as acquisition quality varies.

\subsection{Per-landmark anatomical analysis}\label{subsec:per-landmark}

\begin{figure*}[tp]
\centering
\includegraphics[width=\textwidth,height=0.9\textheight,keepaspectratio]{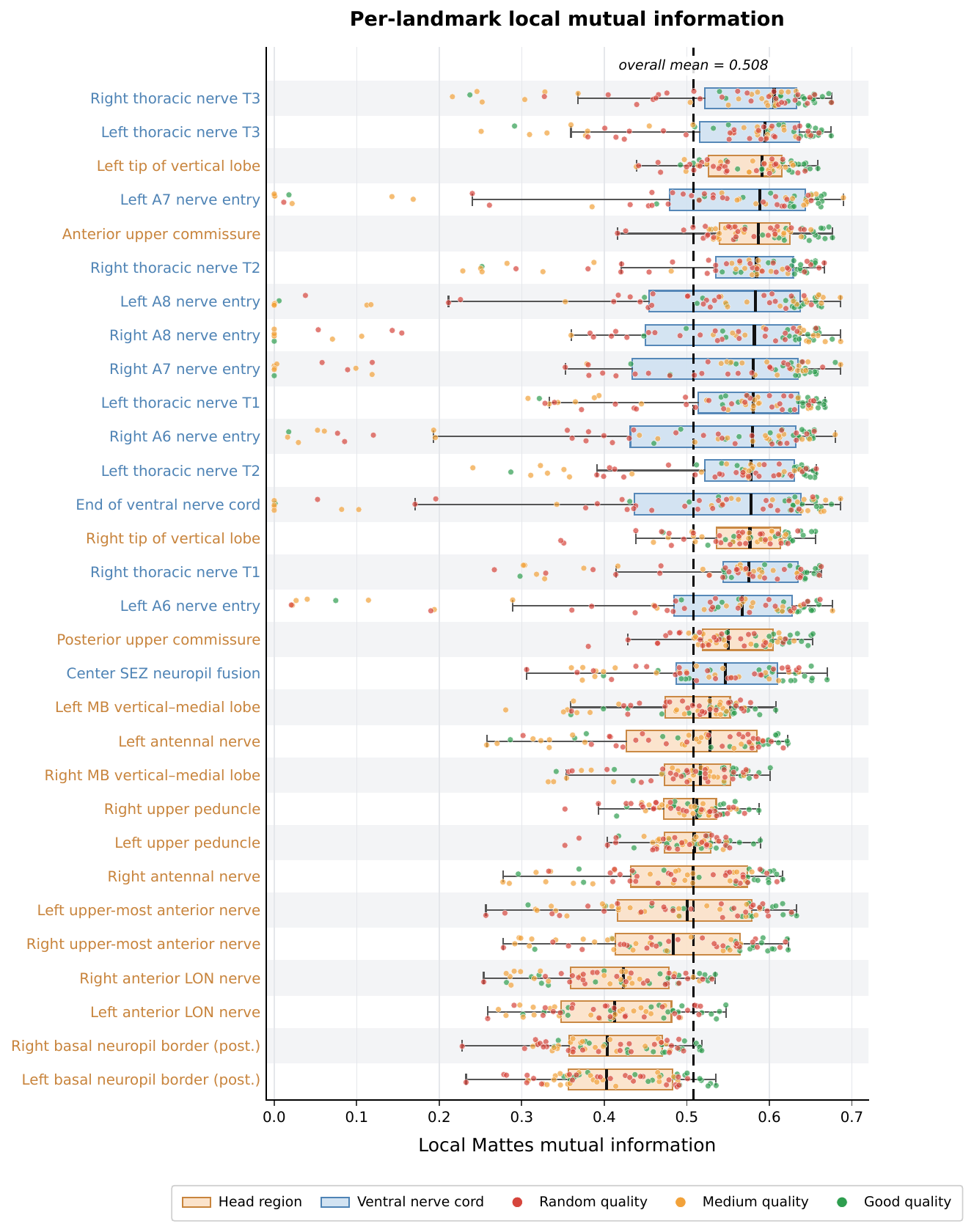}
\caption{\textbf{Per-landmark accuracy of the DLBR model.} Landmark-local
NMI across the 66 Larvalign test cases (cf. Eq.~\ref{eq:nmi-local-mean}). Each row is one of the
$L=30$ landmarks, sorted by median local NMI. Color encodes anatomical region, and each
point is a test case colored by quality tier. The dashed line marks the overall mean.
Higher is better.}
\label{fig:per_landmark_local_mi}
\end{figure*}

Figure~\ref{fig:per_landmark_local_mi} breaks the landmark-local NMI of the best
DLBR model into its $30$ individual landmarks across all test cases. The model
aligns the head structures most accurately, with the commissures, the mushroom-body
lobes and the peduncles at the top of the ranking. The ventral nerve cord and the
paired thoracic and abdominal nerve entries are more challenging, which is expected because these
posterior structures show the largest pose and shape variation across specimens and sit
where the field of view and staining are least consistent. The per-case points confirm
the quality trend of Section~\ref{subsec:robustness}, with good-quality cases
concentrated toward higher local NMI at almost every landmark.

\subsection{Inference speed}\label{subsec:speed}

Registration accuracy is only useful in practice if it comes at an acceptable cost per
volume. The learned model registers one volume in a single forward pass of $g_\theta$,
so the optimization cost is paid once during training rather than separately for every
new brain. Classical methods offer no such amortization, because every new volume
restarts the full iterative optimization from scratch. At the highest resolution the
most accurate classical baseline, diffeomorphic Demons, needs a median of $61.5$ s per
volume, and the elastix affine to B-spline cascade on which the Larvalign pipeline of
\cite{Muenzing2017} is built needs $13.2$ s. We measured that inference time by re-running the
cascade with a current elastix release on current hardware, so that the almost ten years
of processor and software progress since the original publication count in favor of the
baseline rather than inflating our speedup. That publication reported several minutes per
stack for its full pipeline, so we compare against the faster modern re-run rather than
against the published number. DLBR registers the same volume in $0.43$ s
on a GPU, a $144$-fold speedup over Demons and a $31$-fold speedup over the elastix
cascade. The speedup is not traded against quality, since DLBR is at the same time the
most accurate method in Section~\ref{subsec:accuracy}. Sub-second inference is a property
of the learned family as a whole rather than of DLBR alone, so the runtime differences
among the learned methods carry little practical weight and the comparison within that
family rests on accuracy. Larvalign is the reference
pipeline for this species and developmental stage, so this places DLBR one to two orders
of magnitude ahead of the current state of the art for larval brain registration and
turns a per-volume cost measured in tens of seconds into an interactive one.

\subsection{Qualitative results}\label{subsec:qualitative}

Figure~\ref{fig:qualitative_overlay} shows the model on six representative test cases
spanning the quality tiers. For each case the panels show the raw input resampled onto
the reference grid, the preprocessed and affine pre-aligned volume the model consumes,
the predicted deformation field, and the overlay of the moved volume on the fixed
template. The overlays show substantial agreement (indicated in yellow) between the moved brain and
the template, also for the random-quality cases, while the predicted deformation fields remain smooth without obvious local distortions. The visual result matches the quantitative ranking, with the good-quality
cases showing the closest overlays and the random-quality cases still well aligned over
the main neuropil.

\subsection{Ablation studies}\label{subsec:ablation}

We report two ablations that support the main result. The first checks that the headline
high-resolution comparison is not specific to one scale. The second checks that the
featured DLBR configuration is a sound and stable choice within other configurations.

\subsubsection{Resolution-scale ablation}\label{subsubsec:scale-ablation}

\begin{figure*}[t]
\centering
\includegraphics[width=\textwidth]{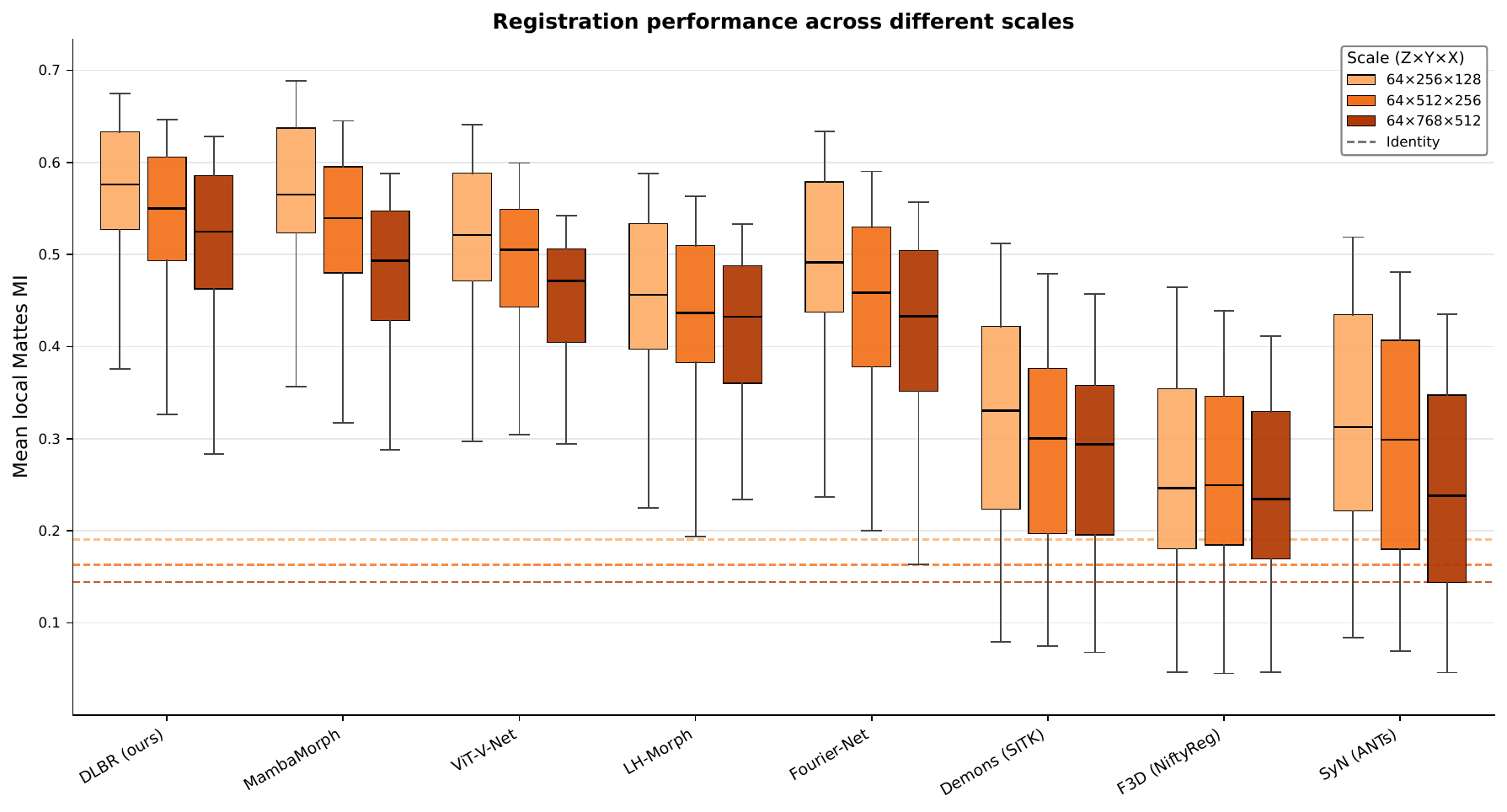}
\caption{\textbf{Registration accuracy across the three preprocessing scales.} Per-case
mean landmark-local NMI over the 66 Larvalign test cases
(cf. Eq.~\ref{eq:nmi-local-mean}) for the five strongest
learned and three strongest classical methods, one box per scale. Dashed lines mark the
no-registration identity floor at each scale. Higher is better.}
\label{fig:scale_ablation_local_mi}
\end{figure*}

Figure~\ref{fig:scale_ablation_local_mi} repeats the comparison at the two coarser
preprocessing scales $64\times256\times128$ and $64\times512\times256$ and places them
next to the featured $64\times768\times512$ scale. The learned methods lead their
classical comparators at every scale, so the advantage of the learned approach is not a
by-product of the chosen resolution. Absolute local NMI tracks the resolution, since
more voxels expose more anatomical detail for the intensity metrics to reward, and the
best DLBR model stays at the top of the learned field across all three scales.
The headline result at $64\times768\times512$ is therefore representative rather than
scale-specific, and the two coarser scales characterize the accuracy that is available
at lower downscaled resolutions.

\subsubsection{Configuration sensitivity}\label{subsubsec:hparam}

\begin{figure}[tp]
\centering
\includegraphics[width=\linewidth,height=0.8\textheight,keepaspectratio]{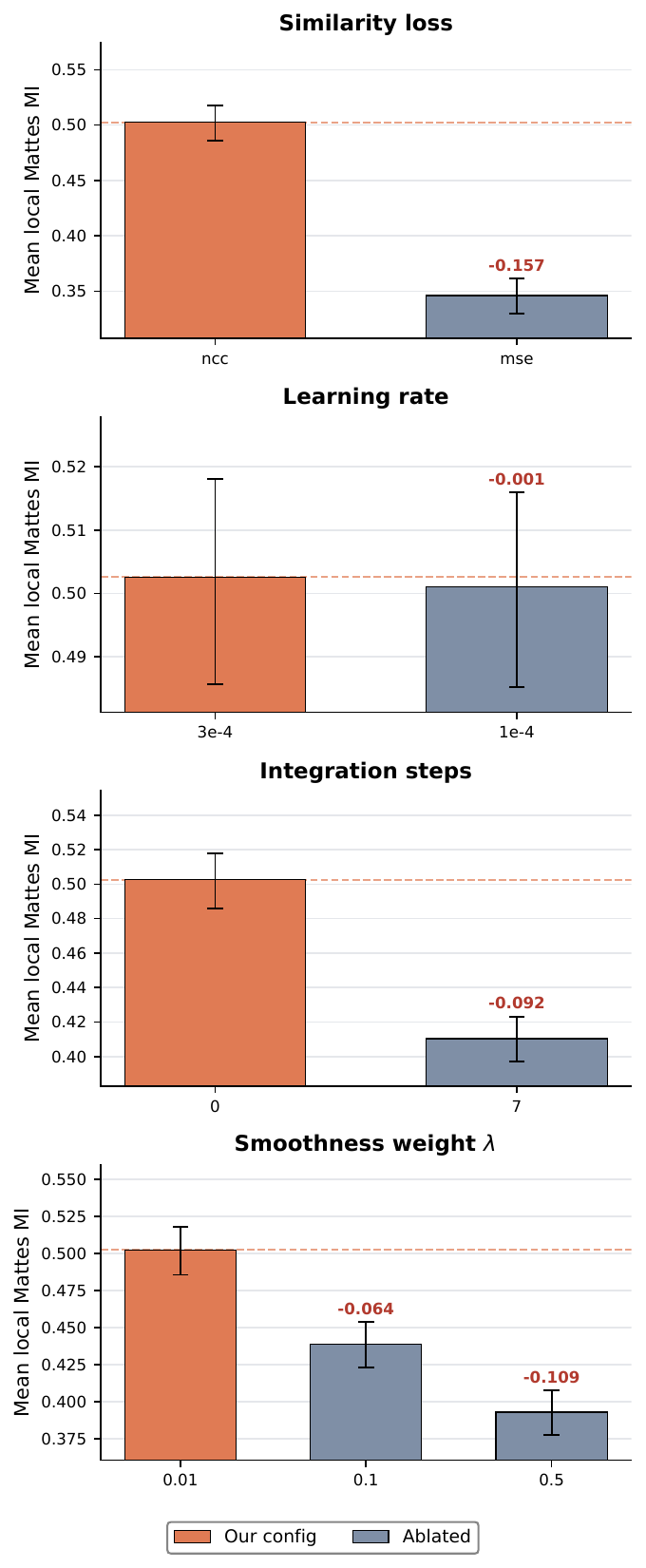}
\caption{\textbf{Configuration ablation of the DLBR training
parameters.} Each panel varies one parameter around the configuration, from
top to bottom the similarity loss, learning rate, integration steps and smoothness weight
$\lambda$. Bars are annotated with the change in mean landmark-local NMI
(cf. Eq.~\ref{eq:nmi-local-mean}) relative to
our configuration in orange, with the dashed line at the best-configuration
accuracy and bootstrap 95 percent confidence intervals over the 66 test cases. Higher is
better.}
\label{fig:voxelmorph_ablation_ofat}
\end{figure}

Figure~\ref{fig:voxelmorph_ablation_ofat} varies the DLBR training
hyperparameters one at a time around the featured configuration, holding all other
settings fixed. We vary the similarity loss term, the learning rate, the number of
integration steps and the smoothness weight $\lambda_{\mathrm{smooth}}$.

The integration steps are the one setting that changes the deformation model itself rather
than a training choice. Instead of the displacement of Eq.~\ref{eq:deformation}, the field
predicted by the network is interpreted as a stationary velocity $v$ and integrated into a
diffeomorphic warp by scaling and squaring, following \cite{dalca2019diffeomorphic}.
Integration starts from a scaled displacement and repeatedly composes the field with
itself,
\begin{equation}
\begin{split}
    \phi^{(0)} &\;=\; \mathrm{Id} + 2^{-N} v, \\
    \phi^{(k+1)} &\;=\; \phi^{(k)}\circ\phi^{(k)},
        \quad k=0,\dots,N-1, \\
    \phi &\;=\; \phi^{(N)},
\end{split}
\label{eq:scaling-squaring}
\end{equation}
where each composition is itself carried out with the spatial transformer of
Eq.~\ref{eq:warp}, and the smoothness penalty of Eq.~\ref{eq:grad} then acts on the
velocity $v$ rather than on a displacement. The number of integration steps $N$ controls
the fidelity of the integration, and setting $N=0$ recovers the plain displacement field
that the featured model uses.

Among the configurations tested, the choice of the similarity term has the largest effect on landmark-local NMI. Replacing normalized cross-correlation by mean squared error costs
$0.157$ of mean landmark-local NMI, by far the largest effect in the ablation. Enabling
scaling-and-squaring integration with seven steps costs $0.092$. Raising the smoothness
weight from $\lambda_{\mathrm{smooth}}=0.01$ to $0.1$ costs $0.064$ and raising it to
$0.5$ costs $0.109$, so an over-regularized field loses the fine local alignment that the
landmark neighborhoods measure. The learning rate is essentially flat, with the two
values separated by $0.001$ of mean landmark-local NMI. The featured configuration is therefore a stable
choice within its neighborhood. The size of the similarity-term effect also shows that a
direct intensity difference by the mean squared error is not an adequate objective for these volumes. Staining and
illumination vary between specimens, so at this resolution mean squared error is driven
by brightness differences rather than by misalignment, whereas normalized
cross-correlation is insensitive to them.

\subsection{Limitations of the trainable resolution}\label{subsec:limitations}

\begin{figure}[t]
\centering
\includegraphics[width=\linewidth]{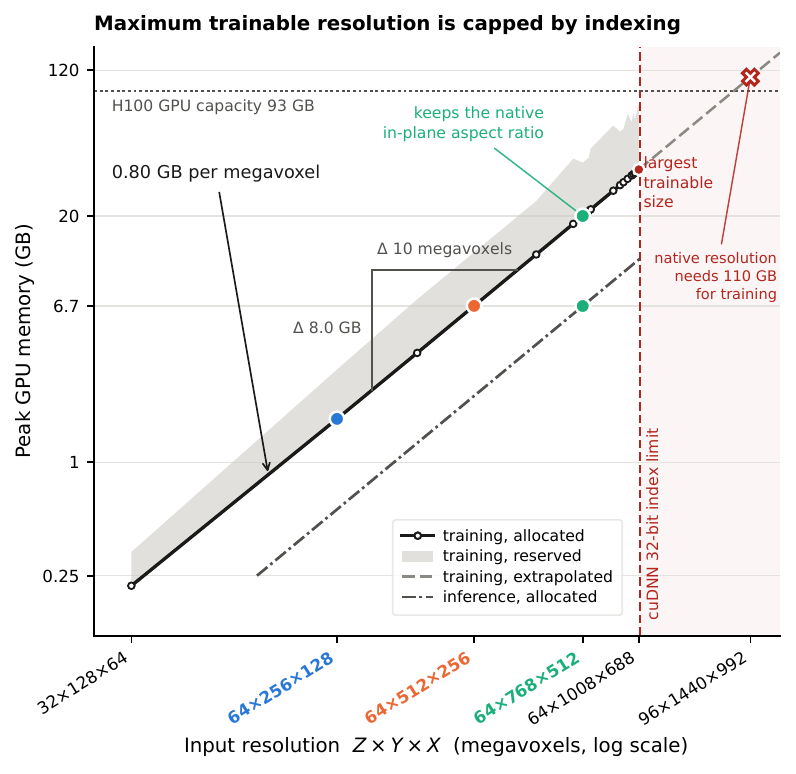}
\caption{\textbf{Maximum input resolution limitations.} Peak GPU memory of a DLBR training
step against input resolution, measured on one NVIDIA H100 with $93$ GB at batch size $1$.
Memory grows at $0.80$ GB per megavoxel, the shaded band is the additional memory the card
must have free, and colored markers are the three preprocessing scales of this study. The
lower curve is the cost at inference. The dashed line marks the resolution above which
training fails for reasons of kernel indexing. Both axes are logarithmic.}
\label{fig:gpu_memory_scaling}
\end{figure}

Finer input grids are desirable in this domain because much of the anatomical signal sits
in structures that are only a few voxels across, such as the nerve entries and the tracts
of the ventral nerve cord. The featured grid resolves these structures, but the maximum training resolution is constrained by the memory required to store activations during training. We measured the peak GPU memory of a full training step across input resolutions on a single NVIDIA H100 with $93$ GB, shown in
Figure~\ref{fig:gpu_memory_scaling}. Memory grows in proportion to the number of voxels at
$0.80$ GB per megavoxel, and almost all of it is held by the full-resolution activations
rather than by the network weights. Training fails altogether above roughly $45$ megavoxels, a limit imposed by
how the kernels index volumes of millions of voxels rather than by the
capacity of the card. Within that bound, $64\times768\times512$ is the resolution we adopt. It keeps the in-plane
aspect ratio at $1.50$, close to the native median of $1.46$, and every axis length remains
divisible by the factor of two applied at each of the three downsampling stages of the
network. The next step in the same series, obtained by doubling the in-plane grid to
$64\times1536\times1024$, contains $101$ megavoxels and is out of reach, as is the native
acquisition grid, which would need about $110$ GB and is moreover still capped by indexing. This limitation applies to training
alone. Registering a new volume with the released model holds no gradients and needs about
$6.7$ GB at the featured resolution, which a modern workstation GPU provides comfortably,
so the hardware demand falls on training the model rather than on applying it.

\section{Discussion}\label{sec:discussion}

The central aim of this work is to provide a registration method for Drosophila larval brain volumes
that is accurate at high spatial resolution and fast
enough to serve as a routine preprocessing step. DLBR meets both requirements. Trained
once on the Janelia collection and applied unchanged to the Larvalign
test dataset, it reaches the best landmark-local NMI, the best global NMI and the best global
NCC of any method we evaluated at the featured $64\times768\times512$ resolution. It provides a clear improvement over the strongest classical baseline in landmark-local NMI, offers a gain of about 23 percentage points over diffeomorphic Demons, and it produces each alignment in a single forward pass of $0.43$ s rather than up to minutes of per-case optimization that the classical pipelines require. The paired tests of
Section~\ref{subsec:robustness} place that margin inside every acquisition-quality
stratum rather than only on the easiest cases, and the resolution-scale ablation shows
that it also holds at the two coarser preprocessing scales.

The results suggest that the distinction between
learned and classical registration is more
consequential than the differences among the
learned architectures evaluated here. On the complete test set every learned network exceeds every classical baseline on both mutual-information scores, and the separation between the two families is several times
larger than the spread inside the learned family. Among the classical methods the
deformable ones lead with diffeomorphic Demons, while the rigid and affine baselines stay close to the floor of the identity baseline. That floor is itself informative, since it shows that what remains after the preprocessing affine is a non-linear error and cannot be recovered by a better cascading linear fit.

The robustness analysis suggests why the learned family separates so cleanly. Classical
registration solves every pair on its own terms, so its similarity metric has to carry all
of the anatomical knowledge, and the staining and illumination differences between
specimens act directly on the objective being optimized. A network trained on $540$ Janelia volumes can instead exploit patterns of anatomical and
acquisition-related variation learned from the training population when processing a new case. This shows
up as a much smaller loss of accuracy when input quality degrades. Going from good to
random quality DLBR loses $10.9$ percent of its landmark-local NMI, the smallest loss of
any method we evaluated, whereas the classical deformable methods lose between $30$ and
$42$ percent. Since the entire Larvalign collection lies outside the training
dataset, the same result is evidence that a model trained in one dataset transfers
to volumes acquired in other settings.

The configuration ablation points in the same direction. The similarity term is by far the
most consequential choice in the DLBR recipe. Replacing normalized cross-correlation by
mean squared error costs around $16$ percentage points of landmark-local NMI, which is larger than the accuracy
difference between any two learned architectures we compared. Mean squared error reads
brightness differences between specimens as misalignment, and that is precisely the
variation this data carries. Two further observations are worth recording. Diffeomorphic
integration by scaling and squaring cost $0.092$ of landmark-local NMI here, so the
featured model uses the predicted field directly as a displacement, and raising the
smoothness weight above $\lambda_{\mathrm{smooth}}=0.01$ cost accuracy at every value we
tried. The network itself is small at $604{,}195$ parameters, so what constrains training
at this resolution is the memory held by full-resolution activations rather than model
capacity, which is the bound Section~\ref{subsec:limitations} quantifies. Registering a
new volume carries none of that cost.

These findings place DLBR in the same role for this domain that the Larvalign pipeline of
\cite{Muenzing2017} played with classical registration. Where Larvalign established an
elastix-based cascade as an accurate way to bring individual larval brains into a shared
anatomical space, DLBR is a learned counterpart that moves the compute cost to a one-time
training stage, removes the per-volume parameter tuning, reaches higher accuracy and stays
the most robust of all methods we compare as acquisition quality degrades.
Together with the released package, registration changes from a step that each study has
to assemble and tune into one that a study can simply call.

\section{Conclusion}\label{sec:conclusion}

We present DLBR, a fast and accurate learned registration network for
Drosophila larval brains that operates at high spatial resolution, leads eleven
classical and seven further learned methods on the
landmark-local accuracy of an external test set, and degrades least when acquisition
quality varies. The network, its trained weights and the full preprocessing and evaluation
pipeline are released as the open-source deep-larval-brain-reg framework to support
reproducible research and adoption in downstream larval-brain analysis.

\backmatter

\bmhead{Acknowledgements}

This work was supported by the German Research Foundation (Deutsche Forschungsgemeinschaft, DFG) under project number 441181781. We thank James W. Truman from the Janelia Research Campus (US) for providing data. The authors gratefully acknowledge the computational and data resources provided by the Leibniz Supercomputing Centre (\url{www.lrz.de}).


\bibliography{sn-bibliography}

\end{document}